\pdfoutput=1
\documentclass{article}
\usepackage{iclr2024_conference,times}

\usepackage{amsmath,amsfonts,bm}

\def\eqref#1{equation~\ref{#1}}

\def\1{\bm{1}}

\DeclareMathAlphabet{\mathsfit}{\encodingdefault}{\sfdefault}{m}{sl}
\SetMathAlphabet{\mathsfit}{bold}{\encodingdefault}{\sfdefault}{bx}{n}

\usepackage{url}            
\usepackage{booktabs}       
\usepackage{amsfonts}       
\usepackage{nicefrac}       
\usepackage{microtype}      
\usepackage{hyperref}       
\usepackage{xcolor}         

\usepackage{amsmath,amsfonts,bm}  
\usepackage{pifont}               
\usepackage{graphicx}             
\usepackage{diagbox}              
\usepackage{multirow}             
\usepackage{enumitem}             
\usepackage{siunitx}              
\usepackage[normalem]{ulem}
\usepackage{wrapfig}
\usepackage{tabularx}
\usepackage{subcaption}
\usepackage{tabu}
\usepackage{mathtools}
\usepackage[misc]{ifsym}
\usepackage[vlined,linesnumbered,ruled]{algorithm2e}

\hypersetup{
  colorlinks,
  linkcolor={red!80!black},
  citecolor={blue!80!black},
  urlcolor={green!50!black}
}

\title{Reuse and Diffuse: Iterative Denoising for Text-to-Video Generation}

\author{
\bfseries
Jiaxi~Gu\textsuperscript{1} \quad Shicong~Wang\textsuperscript{2} \quad Haoyu~Zhao\textsuperscript{2} \quad Tianyi~Lu\textsuperscript{2} \quad Xing~Zhang\textsuperscript{2} \quad Zuxuan~Wu\textsuperscript{2\,\Letter}\\
\bfseries
Songcen~Xu\textsuperscript{1} \quad Wei~Zhang\textsuperscript{1} \quad Yu-Gang~Jiang\textsuperscript{2} \quad Hang~Xu\textsuperscript{1\,\Letter}\\
  \mdseries
  \textsuperscript{1}Huawei Noah's Ark Lab \qquad \textsuperscript{2}Fudan University\\
\Letter\,\texttt{chromexbjxh@gmail.com} \qquad \Letter\,\texttt{zxwu@fudan.edu.cn}
}

\iclrfinalcopy 
\begin{document}

\maketitle

\begin{center}
  \captionsetup{type=figure}
  \includegraphics[width=\linewidth]{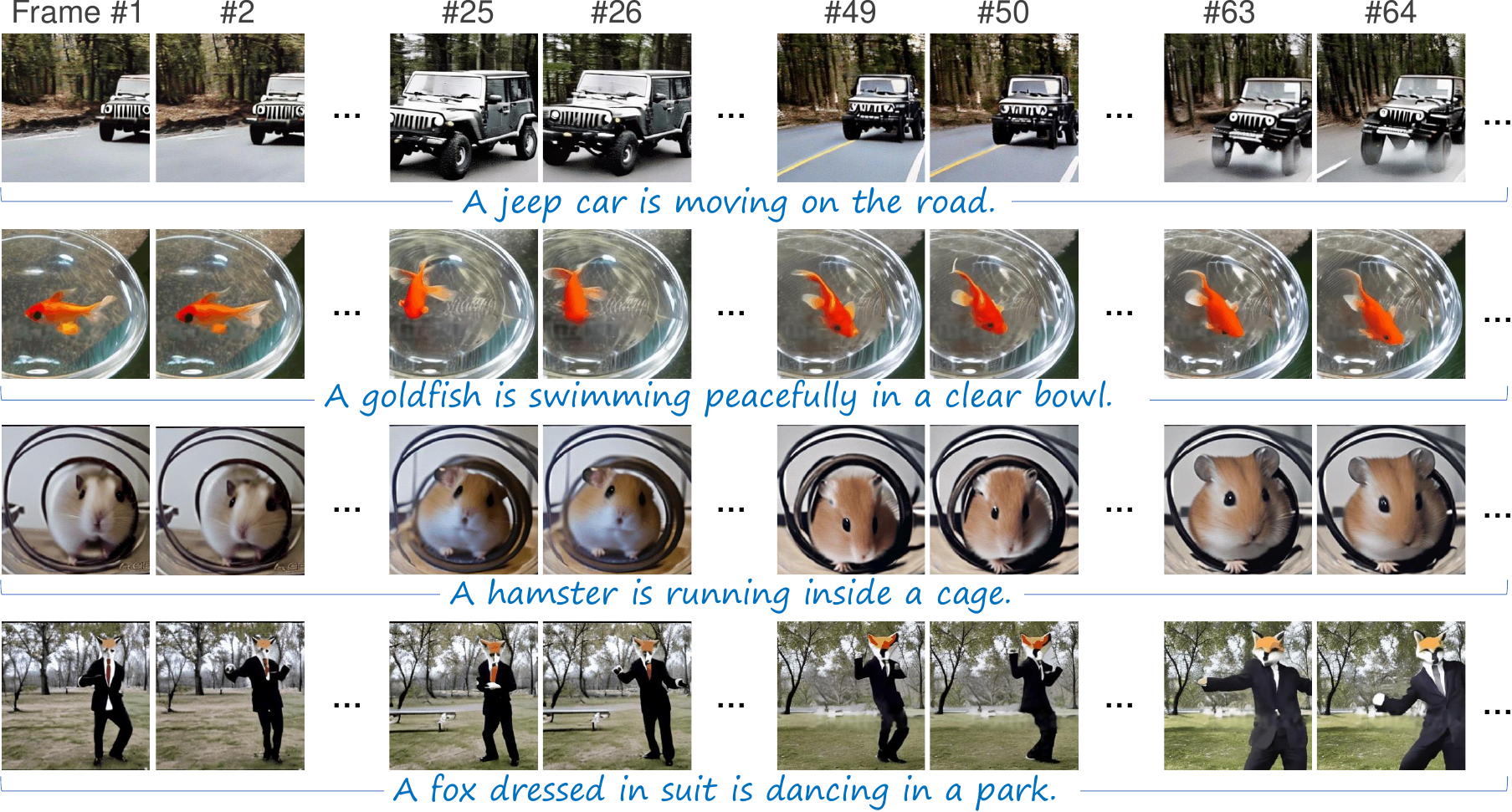}
  \captionof{figure}{These are example videos with diverse content, generated using VidRD, guided by the text prompts below. With a diffusion model for video synthesis, video frames can be generated iteratively by reusing noise and imitating the diffusion process clip by clip. A large number of frames can be finally generated, and smoothness across frames can also be maintained.}
  \label{fig:teaser}
\end{center}

\begin{abstract}

Inspired by the remarkable success of Latent Diffusion Models (LDMs) for image synthesis, we study LDM for text-to-video generation, which is a formidable challenge due to the computational and memory constraints during both model training and inference. A single LDM is usually only capable of generating a very limited number of video frames. Some existing works focus on separate prediction models for generating more video frames, which suffer from additional training cost and frame-level jittering, however. In this paper, we propose a framework called ``Reuse and Diffuse'' dubbed \textit{VidRD} to produce more frames following the frames already generated by an LDM. Conditioned on an initial video clip with a small number of frames, additional frames are iteratively generated by reusing the original latent features and following the previous diffusion process. Besides, for the autoencoder used for translation between pixel space and latent space, we inject temporal layers into its decoder and fine-tune these layers for higher temporal consistency. We also propose a set of strategies for composing video-text data that involve diverse content from multiple existing datasets including video datasets for action recognition and image-text datasets. Extensive experiments show that our method achieves good results in both quantitative and qualitative evaluations. Our project page is available \href{https://anonymous0x233.github.io/ReuseAndDiffuse/}{here}.
\end{abstract}

\section{Introduction}
\label{sec:introduction}

Text-to-video synthesis~\citep{esser2023structure, blattmann2023align, ge2023pyoco} recently has become an increasingly popular research topic in the field of Artificial Intelligence Generated Content (AIGC) following the success of Diffusion Models for image synthesis~\citep{rombach2022high}.
This technique allows businesses to create engaging videos from written text without the need for expensive equipment or professional illustrators. Video creation will become more efficient and innovative with the advancement of artificial intelligence technologies.

Existing video synthesis methods have achieved some progress, but the quality of the generated videos remains less than satisfactory. On the one hand, keeping temporal consistency while generating diverse content remains a big challenge. On the other hand, a typical LDM is only capable of generating a few video frames due to the limitation of computation and memory resources. High-quality smooth videos, involving diverse content and containing a quantity of frames, are preferable in real applications. For this purpose, previous works such as FDM~\citep{harvey2022flexible}, MCVD~\citep{voleti2022mcvd} and Video LDM~\citep{blattmann2023align}, exploit prediction mechanisms for producing future frames based on current video frames. Frame prediction, however, is no easier than direct video generation, and solving frame-level jittering is difficult. Furthermore, a cascaded pipeline, involving a video generation module and a prediction module, introduces more training cost and inference time. In this paper, instead of exploiting frame prediction, we propose a novel framework called ``Reuse and Diffuse'' dubbed \textit{VidRD} and Figure~\ref{fig:teaser} shows some examples generated by it. VidRD can generate more coherent and consistent video frames by leveraging the previous frames generated by a single LDM. After generating an initial video clip by LDM, the following frames are produced iteratively by reusing the latent features of the previous clip and imitating the previous diffusion process. VidRD contains a temporal-aware LDM based on a pre-trained LDM for image synthesis. To train our model efficiently, we initialize the parameters of the spatial layers in the pre-trained image LDM. We also reform and fine-tune the decoder of autoencoder by injecting temporal layers into it. For iterative generation, VidRD contains three novel modules: Frame-level Noise Reversion (FNR), Past-dependent Noise Sampling (PNS), and Denoising with Staged Guidance (DSG). FNR reuses the initial noise in reverse order from the previous video clip, while PNS brings a new random noise for the last several video frames. Furthermore, temporal consistencies between video clips are refined by DSG.

Moreover, the training of LDMs usually relies on a massive amount of data to ensure the quality of the generative content~\citep{khachatryan2023text2video}. The scarcity of high-quality video-text data has always been a problem. To this end, we devise a set of strategies to utilize existing datasets including video datasets for action recognition and image-text datasets. In addition to the typical video datasets in which each video is captioned with a short descriptive sentence, we use multi-modal Large Language Models (LLMs) to segment and caption videos in action recognition video datasets. Additionally, images with text captions are transformed to pseudo-videos by random zooming and panning so visual content of videos can be largely enriched.

Extensive experiments demonstrate that VidRD consistently achieves high performance in both quantitative and qualitative evaluations. On the benchmark of video generation based on UCF-101~\citep{soomro2012ucf101}, we achieve Fr\'echet Video Distance (FVD) of $363.19$ and Inception Score (IS) of $39.37$.


In summary, our contributions are three-fold:
\begin{itemize}
    \item We propose VidRD, an iterative text-to-video generation method that leverages a temporal-aware LDM to generate smooth videos. By reusing the latent features of the initially generated video clip and imitating the previous diffusion process each time, the following video frames can be produced iteratively.
    \item A set of effective strategies is proposed to compose a high-quality video-text dataset. We use LLMs to segment and caption videos from action recognition datasets. Image-text datasets are also used by transforming into pseudo-videos with random zooming and panning.
    \item Extensive experiments on the UCF-101 benchmark demonstrate that VidRD achieves good FVD and IS in comparison with existing methods. Qualitative evaluations also show good results.
\end{itemize}

\section{Related Work}
\label{sec:related_work}

\paragraph{Image synthesis models.}

Automatic image synthesis is seen as a major milestone towards general artificial intelligence~\citep{goertzel2007artificial, clune2019aigas, fjelland2020general, zhang2023texttoimage}. In the early stage, GAN (Generative Adversarial Network)-based~\citep{xu2018attngan, li2019controllable, karras2019style} and autoregressive methods~\citep{ramesh2021zeroshot, ding2021cogview, chang2022maskgit, wu2022nuwa, yu2022scaling} are designed to achieve image synthesis. DALL-E~\citep{ramesh2021zeroshot} and Parti~\citep{yu2022scaling} are two representative methods. DALL-E proposes a two-stage training process: image tokens are generated and then combined with text tokens in the second stage. However, the autoregressive nature of the above methods results in high computation costs and sequential error accumulation~\citep{zhang2023texttoimage}. Text-guided image synthesis afterward makes significant progress due to the advancement of Diffusion Models (DMs)~\citep{rombach2022high}. Due to the surprising results of DMs, massive works~\citep{kawar2023enhancing, bhunia2023person, liu2023more, fan2023frido, dabral2023mofusion, huang2023composer, ramech2022hierarchical, saharia2022photorealistic} focus on this technology. GLIDE~\citep{nichol2021glide} and Stable Diffusion~\citep{rombach2022high} are two representative DM-based works that employ Vision-Language Models (VLMs) such as CLIP~\citep{radford2021learning} for text-guided image synthesis. With the rapid development of DMs, advanced image editing task has also been achieved.

\paragraph{Video synthesis models.}

Through the lens of AIGC, text-guided video synthesis has also attracted massive attention after image synthesis. Earlier methods focus on unconditional video synthesis. Similar to image synthesis, two kinds of approaches are proposed including GAN-based~\citep{vondrick2016generating, saito2017temporal, acharya2018towards, tulyakov2018mocogan} and transformer-based~\citep{ge2022long, yan2021videogpt, srivastava2015unsupervised, moing2021ccvs, hong2023cogvideo}. StyleGAN-V~\citep{skorokhodov2022styleganv} uses time-continuous signals for video synthesis. They extend the paradigm of neural representations to build a continuous-time video generator. In VideoGPT~\citep{yan2021videogpt}, VQ-VAE~\citep{van2016pixel} is used to learn subsampled discrete latent representations of raw videos. The latent representations are modeled using a robust autoregressive prior with a GPT-like architecture. However, GAN-based methods are difficult to scale up to handle complex and diverse video distributions due to their susceptibility to mode collapse. Also, they suffer from training instability issues. Recently, motivated by DM-based image synthesis, several works~\citep{esser2023structure, ge2023pyoco, blattmann2023align, khachatryan2023text2video, he2022latent, dabral2023mofusion, luo2023videofusion, brooks2022generating} propose to explore DMs for conditional video synthesis. Among these works, Video LDM~\citep{blattmann2023align} is a representative work and also exhibits excellent results. On the basis of an LDM for image synthesis pre-trained on large-scale image-text data, Video LDM fine-tunes its newly added temporal layers with video data. In addition, the authors propose an interpolation model and an upsampler model for generating high-quality videos. Almost at the same time, PYoCo~\citep{ge2023pyoco} is proposed as an improved method for extending LDM from image synthesis to video synthesis. Based on the continuity of video content over time, the authors design a video noise prior to achieve better temporal consistency. With the satisfactory results of LDM-based video synthesis, there are also some works~\citep{molad2023dreamix, qi2023fatezero, liu2023videop2p} on controllable video editing.

\paragraph{Iterative video generation.}

Current video synthesis methods can only deal with several sparsely sampled video frames limited by computational and memory resources. Longer video synthesis, even for a single scene, is still a challenging topic. TATS~\citep{ge2022long} proposed a method building on 3D-VQGAN with transformers~\citep{vaswani2017attention} to generate videos with thousands of frames. With the rise of DM, Video LDM~\citep{blattmann2023align} designed a separate LDM as a prediction model for extending the already generated video clip. This multiplies the training cost for video synthesis. Also in latent space, LVDM~\citep{he2022latent} proposed a unified model by treating frame interpolation and prediction as two sub-tasks in model training. This makes training more difficult and the training process unstable. In this work, we aim to build a unified and efficient text-to-video generation model for generating high-quality smooth videos containing a large number of frames.

\section{Preliminaries}
\label{sec:preliminaries}

\subsection{Latent Diffusion Models}

DMs learn to model a data distribution $p_\text{data}$ via iterative denoising from a noise distribution so the desired data distribution can be generated. Given samples $\mathbf{x_0} \sim p_\text{data}$, the diffusion forward process iteratively adds noise:
\begin{equation}
\label{eq:addnoise_one_step}
    q(\mathbf{x_t} \mid \mathbf{x}_{t-1}) = \mathcal{N}(\mathbf{x}_t; \alpha_t \mathbf{x}_{t-1}, \sigma^2_t\mathbf{I})
\end{equation}
which represents the conditional density of $\mathbf{x}_{t}$ given $\mathbf{x}_{t-1}$. Here, a noise schedule is defined by $\alpha_t$ and $\sigma_t$ parameterized by diffusion time $t$. For generating a fully random noise with the increase of diffusion time $t$, signal-to-noise ratio $\lambda_t = \log(\alpha_t^2 / \sigma_t^2)$ needs to monotonically decrease. To this end, a variance-preserving time schedule satisfying $\alpha^2_t + \sigma^2_t = 1$ is usually used. Following the closure of normal distribution, we can directly sample $\mathbf{x}_t$ at any diffusion time $t$ by:
\begin{equation}
\label{eq:addnoise_from_scratch}
    q(\mathbf{x_t} \mid \mathbf{x}_0) = \mathcal{N}(\mathbf{x}_t; \bar{\alpha}_t \mathbf{x}_0, (1 - \alpha^2_t)\mathbf{I})
\end{equation}
where $\bar{\alpha}_t=\prod_{i=1}^t \alpha_i$.

In the backward process of diffusion, a model denoted by $f_\theta$ parameterized by $\theta$ is trained to predict the noise to iteratively recover $\mathbf{x}_0$ from $\mathbf{x}_T$ which is noisy data after adding noise $T$ times. As long as $T$ is large enough, the original sample of real data is fully perturbed such that $\mathbf{x}_T \sim \mathcal{N}(\mathbf{0}, \mathbf{I})$. While training, the denoising matching score is optimized following:
\begin{equation}
\label{eq:optimization_func}
  \mathbb{E}_{\mathbf{y} \sim \mathcal{N}(\mathbf{0}, \mathbf{I})} [\lVert \mathbf{y} - f_{\theta} (\mathbf{x}_t; \mathbf{c}, t) \rVert_2^2]
\end{equation}
where $\mathbf{y}$ representing the target features can be a random noise and $\mathbf{c}$ is an optional conditioning signal such as text prompt in text-to-something DMs. Also, $t$ is sampled from a uniform distribution which is set to $\mathcal{U}\{0, 1000\}$ in Stable Diffusion~\citep{rombach2022high}. Once $f_\theta$ is trained, we can generate a novel $\mathbf{\mathbf{x}_0}$ from a random noise $\mathbf{x}_T \sim \mathcal{N}(\mathbf{0}, \mathbf{I})$ with a deterministic sampling DDIM~\citep{song2021denoising}.

In practice, since both images and videos rely on large amounts of computational and memory resources, DMs in pixel space are costly. Stable Diffusion~\citep{rombach2022high} proposes to apply a regularized autoencoder to compress the original pixels into latent space to save computation and memory. In this way, DMs are transformed into Latent Diffusion Models (LDMs). The autoencoder in an LDM consists of an encoder $\mathcal{E}$ for encoding pixel features $\mathbf{x}$ into latent features $\mathbf{z} = \mathcal{E}(\mathbf{x})$ and a decoder $\mathcal{D}$ for decoding $\mathbf{z}$ back to $\mathbf{x} = \mathcal{D}(\mathbf{z})$. In general, the autoencoder is trained by reconstructing:
\begin{equation}
  \hat{\mathbf{x}} = \mathcal{D}(\mathcal{E}(\mathbf{x})) \approx \mathbf{x}
\end{equation}
where $\hat{\mathbf{x}}$ denotes the reconstructed sample after the real data $\mathbf{x}$ is input into the encoder and then the decoder in turn. Usually, in addition to the construction loss, an adversarial loss defined by an individual patch-based discriminator is also added. With a trained autoencoder, the latent representations $\mathbf{z}$ after encoding $\mathbf{x}$ with $\mathcal{E}$ can be directly used in Equation~\ref{eq:addnoise_one_step}, \ref{eq:addnoise_from_scratch} and \ref{eq:optimization_func}. After generating $\mathbf{z}_0$ from LDM, the original pixel features $\mathbf{x}_0$ can be restored from latent features with the trained decoder $\mathcal{D}$. For the implementation of an autoencoder of an LDM for image synthesis such as Stable Diffusion, both the encoder $\mathcal{E}$ and the decoder $\mathcal{D}$ are for static images only. For an LDM for video synthesis, the existing autoencoder works frame by frame so no temporal information is considered.

\section{Method}
\label{sec:method}

\subsection{Model Architecture}

\begin{figure}[ht]
    \centering
    \includegraphics[width=\linewidth]{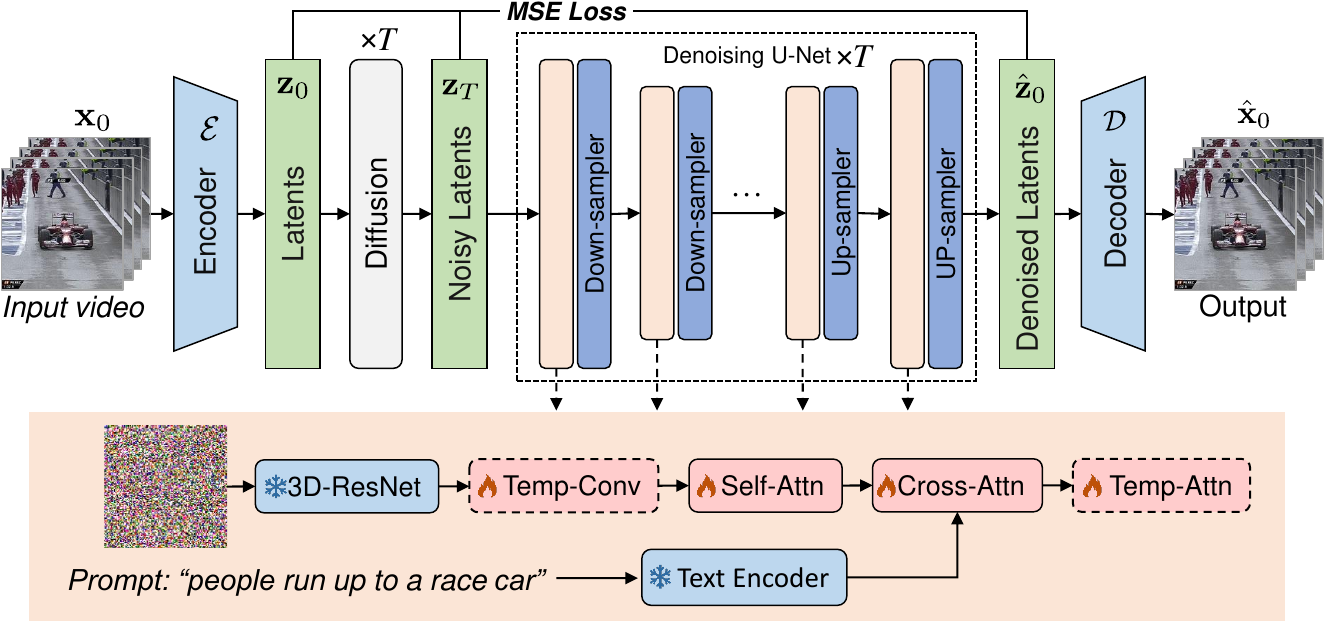}
    \caption{The architecture of VidRD is derived from an LDM for image synthesis. Modules with snowflake marks are frozen while those with flame marks are trainable. Modules with dashed boxes are added in addition to the original LDM for image synthesis.}
    \label{fig:method_arch}
\end{figure}

Since models represented by Stable Diffusion~\citep{rombach2022high} show amazing results in image synthesis, loading its pre-trained LDM is preferable for efficiently training an LDM for high-quality video synthesis~\citep{blattmann2023align,ge2023pyoco}. Similarly, VidRD is based on the pre-trained Stable Diffusion including its Variational Auto-Encoder (VAE) for latent representation and U-Net for latent denoising. Figure~\ref{fig:method_arch} shows the architecture of VidRD. Similar to existing works like Video LDM~\citep{blattmann2023align} and PYoCo~\citep{ge2023pyoco}, we adapt the original U-Net for image diffusion to video synthesis by injecting temporal layers marked with dashed boxes in the figure. These two types of temporal layers are: \textit{Temp-Conv} representing 3D convolution layers and \textit{Temp-Attn} representing temporal attention layers. Also, most network layers, except for the newly added \textit{Temp-Conv} and \textit{Temp-Attn}, in our devised U-Net, are initialized with the pre-trained model weights of Stable Diffusion.  The parameters of \textit{Temp-Conv} and \textit{Temp-Attn} are randomly initialized with the last layer zeroed and residual connections are also applied.

For training this model, videos with text captions are input and then encoded by a text encoder to $\mathbf{c}$ in Equation~\ref{eq:optimization_func} for conditioning in the process of denoising. As inputs, videos are usually sampled at regular intervals into a sequence containing a fixed number of frames. Therefore, we have input videos denoted by $\mathbf{x} \in \mathbb{R}^{B \times F \times 3 \times H^{\prime} \times W^{\prime}}$ where $B$, $F$, $H^{\prime}$ and $W^{\prime}$ respectively denote the inputs' batch size, number of frames, height, and width in the pixel space. After encoding, we can get its corresponding representation in the latent space $\mathbf{z} = \mathcal{E}(\mathbf{x}) \in \mathbb{R}^{B \times F \times C \times H \times W}$ where $C$, $H$ and $W$ respectively denote their channel, height, and width in the latent space.

For dealing with video inputs, the original 2D ResNet of Stable Diffusion is inflated to \textit{3D-ResNet} by fusing the inputs' temporal dimension into batch dimension. In this way, this part of network parameters can be directly inherited from Stable Diffusion. For an input $\mathbf{z} = \mathcal{E}(\mathbf{x}) \in \mathbb{R}^{B \times F \times C \times H \times W}$, it is transformed into $\mathbb{R}^{(BF) \times C \times H \times W}$.

\begin{figure}[ht]
    \centering
    \includegraphics[width=.7\linewidth]{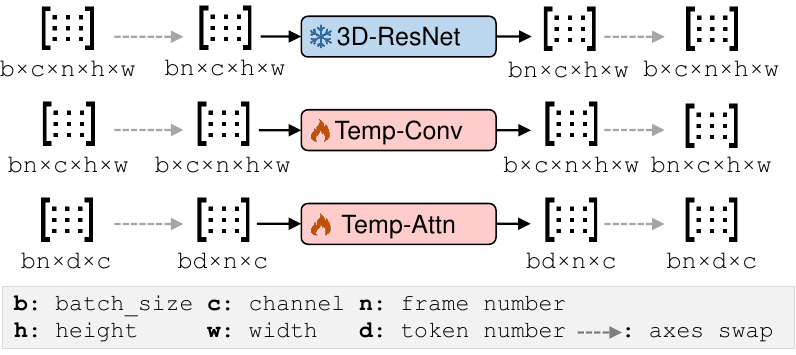}
    \caption{These are three essential network layers in VidRD.  \textit{3D-ResNet}, inherited from Stable Diffusion, treats the number of frames $n$ as a part of batch size. This is equivalent to applying the original \textit{2D-ResNet} frame by frame so this layer is frozen in model training. \textit{Temp-Conv}, implemented with 3D convolutions, processes video inputs in a tube manner while \textit{Temp-Attn} applies attention layer along temporal axis.}
    \label{fig:method_modules}
\end{figure}

For strengthening temporal relations between video frames, two temporal modules are added. One temporal module is \textit{Temp-Conv} implemented with 3D convolution layers which are added right after the ResNet block. Another temporal module is \textit{Temp-Attn} which is an attention layer similar to \textit{Self-Attn} in the original Stable Diffusion, but applied on the temporal dimension. Specifically, for \textit{3D-ResNet}, \textit{Temp-Conv} and \textit{Temp-Attn}, the axes of input data are swapped accordingly to make the whole model work. The implementation details are shown in Figure~\ref{fig:method_modules}.

Similar to existing works like Video LDM~\citep{blattmann2023align} and PYoCo~\citep{ge2023pyoco}, only part of our network layers are trainable for efficient training. For the text encoder, all the parameters are frozen. For U-Net, existing works use either a two-stage~\citep{blattmann2023align} or alternating~\citep{ge2023pyoco} training scheme with image and video data. Essentially, they use image data for fine-tuning spatial layers and video data for training temporal layers, respectively. Instead of this manual training scheme, our U-Net is trained with pure video data in an end-to-end unified way since the image data are transformed into pseudo-videos that show temporal consistency like the original video data. Specifically, as illustrated in Figure~\ref{fig:method_modules}, the network modules in U-Net with red background, including two newly added temporal layers and the spatial attention layers originally designed in LDM for image synthesis, are trainable. Furthermore, since the autoencoder is originally designed for image synthesis, temporal relations between video frames are not considered. To achieve a more accurate representation of videos in the output pixel space, we inject temporal layers implemented with 3D convolutions into the decoder $\mathcal{D}$ and fine-tune it with video data. The details are presented in Section~\ref{subsec:vae_ft}.

\subsection{Video-text data composition}
\label{sec:data_compose}

\begin{figure}[ht]
    \centering
    \includegraphics[width=\linewidth]{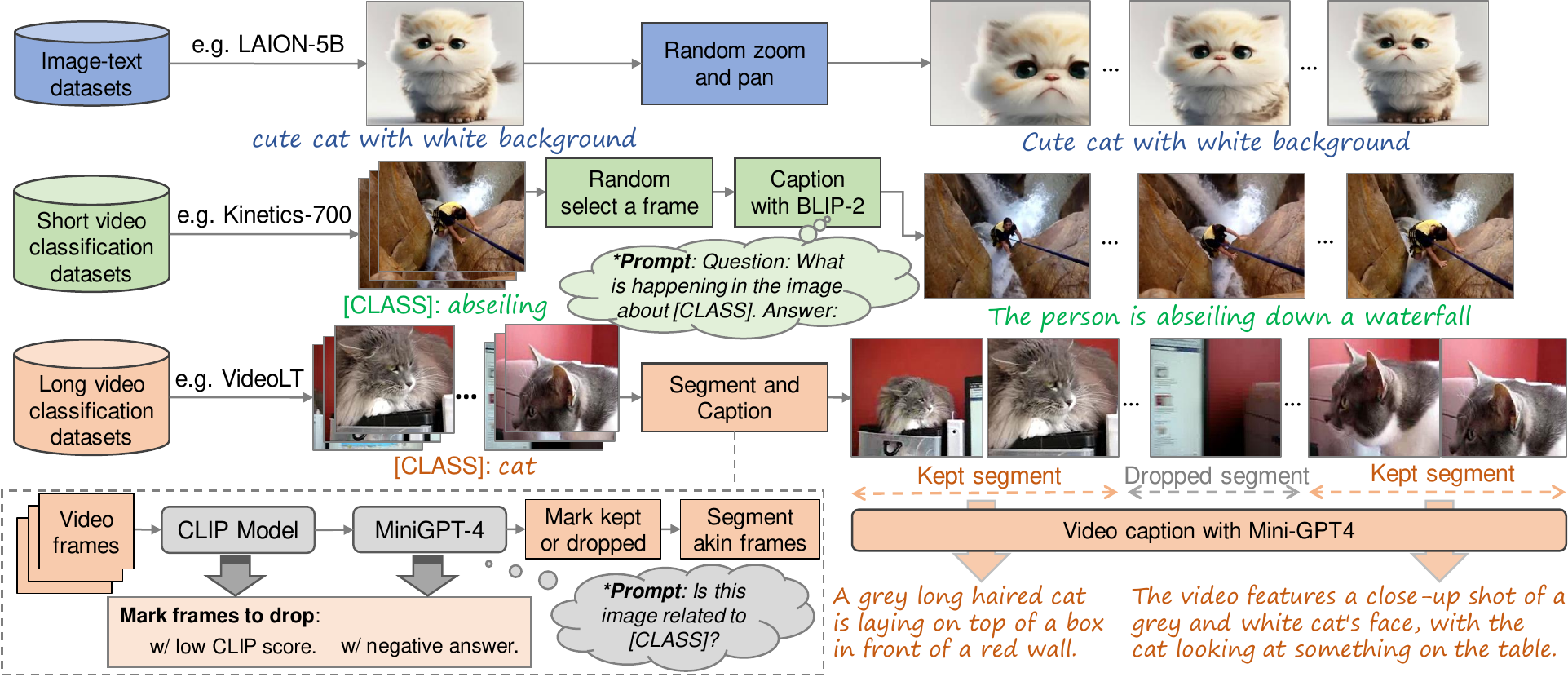}
    \caption{A set of strategies is devised for processing different types of datasets including image-text datasets, short video classification datasets, and long video classification datasets.}
    \label{fig:method_data}
\end{figure}

The training of an LDM for text-guided video synthesis requires a large amount of captioned videos. Large-scale well-captioned video datasets such as VATEX~\citep{wang2019vatex} or WebVid-2M~\citep{bain2021frozen} are very limited however. To compensate for the lack of high-quality video-text data, we propose a set of strategies for composing video-text data from different types of existing datasets other than well-captioned video-text datasets like VATEX~\citep{wang2019vatex} and WebVid-2M~\citep{bain2021frozen}. Figure~\ref{fig:method_data} illustrates the three types of datasets that we use: text-image datasets, short video classification datasets, and long video classification datasets.

\paragraph{Image-text datasets.} Since the quantity and quality of image datasets are much better than those of video datasets, it is necessary to exploit image datasets to improve the video generation model. Also, existing works~\citep{ho2022imagenvideo,wu2022nuwa,ge2023pyoco,blattmann2023align} have revealed that it is beneficial to train a video generation model with image-text data for better appearances of video content. Training spatial network layers with static images is of little help for improving temporal consistency, however. To this end, we propose a better strategy using widely available image-text datasets. As shown in the top part of Figure~\ref{fig:method_data}, for each image with a caption, we apply random zooming and panning to produce multiple images and they are further composed into a pseudo-video. FFmpeg~\citep{tomar2006converting} is used for the implementation of random zooming and panning.

\paragraph{Short video classification datasets.} For those video datasets mostly containing short videos originally used for action recognition such as Kinetics-700~\citep{smaira2020short}, the problem is how to give a proper text caption based on its classification label to each video. For this problem, we employ an LLM with visual comprehension capabilities: BLIP-2~\citep{li2023blip}. Since short videos usually involve a single scene, it is reasonable to use the caption for a randomly selected video frame as the caption for the whole video. This strategy is illustrated as green modules in Figure~\ref{fig:method_data}. For each short video with a given classification label such as ``abseiling'', we randomly select a video frame from it and then use BLIP-2 to generate a text caption by querying the LLM with the frame and its classification label. In order to make the LLM produce more diverse text captions, we use a few prompt templates for querying the LLM, and one of them is randomly selected every time. Due to the length of the paper, only one of the prompt templates is shown in the figure, and the whole list is put in the supplementary material.

\paragraph{Long video classification datasets.} There is also another type of video dataset for action recognition mostly containing long videos involving multiple scenes such as VideoLT~\citep{zhang2021videolt}. For these videos, it is improper to use the description of a single frame as the caption for the whole video. Also, long videos usually contain some irrelevant content that is unrelated to its classification label and not worth captioning. To extract well-caption videos from these long videos, we use a segment-then-caption strategy as shown in the bottom part of Figure~\ref{fig:method_data}. For each video containing multiple frames, we employ CLIP~\citep{radford2021learning} with vision-language alignment ability and MiniGPT-4~\citep{zhu2023minigpt}, an LLM with vision-language understanding to mark those frames irrelevant to the classification label. Those frames with low CLIP matching scores with classification labels or considered irrelevant by MiniGPT-4, while querying it with a devised prompt template, are marked to be dropped. In this way, each frame in a video is marked as kept or dropped. A video can then be segmented with this attribute of each frame. To avoid producing too short videos, we drop those segments with too few frames. Finally, MiniGPT-4 is again used for captioning the segmented sub-videos with devised prompt templates.

\subsection{Longer video generation}

\begin{figure}[ht]
    \centering
    \includegraphics[width=.75\linewidth]{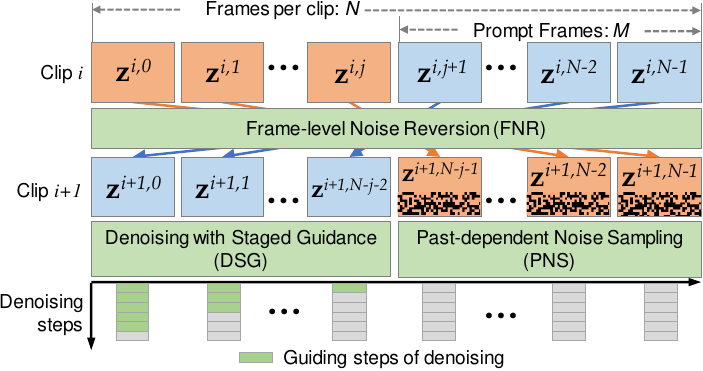}
    \caption{Videos can be generated clip by clip iteratively with a single LDM. After each iteration, $N$ frames are generated and the last $M$ frames are used as prompt frames for the next iteration. Three key strategies are proposed for generating natural and smooth videos. Frame-level Noise Reversion (FNR) is used as a basic module for re-using the initial noise in a reversed order from the last video clip. Past-dependent Noise Sampling (PNS) brings new random noise for the last several video frames. Temporal consistencies between video clips are refined by Denoising with Staged Guidance (DSG).}
    \label{fig:method_iterative}
\end{figure}

A since LDM for video synthesis requires a large amount of computation and memory, it is usually unable to generate a lot of frames at once. Previous works like MCVD~\citep{voleti2022mcvd} and Video LDM~\citep{blattmann2023align} train a separate prediction model for generating the new video frames following the already generated frames. Unlike these methods implemented by frame prediction, we propose an iterative approach with a single diffusion model without fine-tuning. Figure~\ref{fig:method_iterative} shows the pipeline of VidRD for generating more frames that are natural and smooth along with the existing video frames. Within each iteration, there are $N$ frames generated in total and the last $M$ frames are used as \textit{prompt frames} for the next iteration. For the sake of brevity, we use the latent representation of frames in each video clip and omit the use of VAE and text prompts. We use $\mathbf{z}^{i,j}$ to represent the latent features of the $j$-th frame in the $i$-th video clip. Once the initial video clip numbered $0$ is generated, the $i+1$-th clip can be generated on the basis of the $i$-th one iteratively. In this way, a video containing a number of frames can finally be derived by concatenating the continuous video clips. To this end, three key strategies are proposed including \textit{Frame-level Noise Reversion} (FNR), \textit{Past-dependent Noise Sampling} (PNS), and \textit{Denoising with Staged Guidance} (DSG). The whole process is briefly outlined in Algorithm~\ref{alg:iterative_generation}. The detailed implementations are stated below.

\begin{algorithm}[H]
  \KwIn{$\{\mathbf{z}_t^{0,j} \mid j \in \mathbb{Z} \cap [0, N), t \in \mathbb{Z} \cap [0, T]\}$: Latent features of the first clip.}
  \KwOut{$\{\mathbf{z}_0^{i,j} \mid j \in \mathbb{Z} \cap [0, N), i \in \mathbb{Z} \cap [0, V_\text{max})\}$: Denoised latent features of all $V_\text{max}$ clips.}
  \For{$i=1$ \KwTo $V_\text{max}$}{
    \For{$j=0$ \KwTo $N-1$}{
      $\mathbf{z}_T^{i,j} = \mathbf{z}_T^{i-1,N-j-1}$ \tcp*{Frame-level Noise Reversion (FNR)}
      \If(\tcp*[f]{Past-dependent Noise Sampling (PNS)}){$j \geq M$}{
        $\mathbf{z}_T^{i,j}=\mathbf{z}_T^{i,j}\alpha/\sqrt{1+\alpha^2}+\mathbf{\epsilon}^{i,j}; \;\; \mathbf{\epsilon}^{i,j} \in \mathcal{N}(\mathbf{0}, \mathbf{I}/(1 + \alpha^2))$\;
       }
    }
    \For(\tcp*[f]{Denoising with Staged Guidance (DSG)}){$t=T$ \KwTo $1$}{
      \lIf{$t > (1-\beta)T + \beta T j / M$}{$\mathbf{z}_{t-1}^{i,j} = \mathbf{z}_{t-1}^{i-1,N-j-1}$} \lElse{$\mathbf{z}_{t-1}^{i,j} = \text{DDIM}(\mathbf{z}_t^{i,j}, t)$ \tcp*[f]{Progressively denoising with DDIM}}
    }
  }
  \caption{Iterative video generation.}
  \label{alg:iterative_generation}
\end{algorithm}

\paragraph{Frame-level Noise Reversion.}

As previous works have revealed, for generating smooth videos, the initial noise of LDMs for video synthesis is essential~\citep{ge2023pyoco}, and sharing a base noise across video frames also helps~\citep{luo2023videofusion}. We borrowed a similar scheme for generating longer videos and reused the initial noise iteratively but in a reversed order every time. Specifically, the initial video clip is first generated purely with our trained U-Net model by iterative denoising from an initial noise sampled from a normal distribution like:
\begin{equation}
\label{eq:noise_reverse_zero}
    \mathbf{z}^{0,j}_T \sim \mathcal{N}(\mathbf{0}, \mathbf{I}), \;\; j \in \{0, 1, \dots, N-1\}
\end{equation}
where $N$ is the number of frames of a single video clip. The process of denoising from such random noise follows the common practice of most video LDMs. In this way, we generate the initial video clip containing $N$ frames and intend to generate the subsequent video clip. Since the continuous video clips will ultimately need to be concatenated, the temporal consistency between frames, especially across clips, is essential. We propose FNR for this purpose.
The initial random noises can be used to generate $N$ basic video frames with high temporal consistency so intuitively new temporal-consistent frames can also be generated with these noises but in a reversed order. Therefore, considering the initial noises, we have:
\begin{equation}
\label{eq:noise_reverse_iter}
  \mathbf{z}^{i,j}_T = \mathbf{z}^{i-1, N-j-1}, \;\; i \geq 1,\, j \in \{0, 1, \dots, N-1\}
\end{equation}
In combination with Equation~\ref{eq:noise_reverse_zero}, the initial noise of each frame in the following video clips can be computed. From the perspective of frame order, the frame-level order of the initial noise of each frame in the current clip can be seen as a reversion of the ones from the last clip. In this way, the temporal consistency in the joint of two continuous video clips can be ensured to some extent. However, FNR alone cannot guarantee that videos are natural and smooth and the video content may simply become repetitive within a single clip in some extreme cases. We call this phenomenon ``content cycling''. To this end, we propose PNS and DSG.

\paragraph{Past-dependent Noise Sampling.}

In order to mitigate the extent of video content cycling, which is critical to perception, simply reusing the original random noise repeatedly is not enough and randomness needs to be introduced along with the generation of the video clips following the initial clip. Therefore, we propose Past-dependent Noise Sampling (PNS) which is used for introducing randomness gradually. Specifically, excluding the $M$ prompt frames for prompting the generation of the current video clip, random noises are added to the initial noises of the remaining $N-M$ frames, which are initialized with that of $N-M$ frames of the previous video clip. Therefore, Equation~\ref{eq:noise_reverse_iter} is modified based on the position of the frames in each video clip:
\begin{equation}
\label{eq:past_dep_noise_sampling}
  \mathbf{z}^{i,j}_T =
  \begin{cases}
      \mathbf{z}_T^{i-1, N-j-1} & \text{if } j < M \\
      \frac{\alpha}{\sqrt{1 + \alpha^2}}\mathbf{z}_T^{i-1, N-j-1} + \mathbf{\epsilon}^{i,j} & \text{otherwise}
  \end{cases}, \;\;
   \mathbf{\epsilon}^{i,j} \sim \mathcal{N}(\mathbf{0}, \frac{1}{1 + \alpha^2}\mathbf{I}), \; \alpha \geq 0
\end{equation}
where $\mathbf{\epsilon}^{i,j}$ is a newly added random noise, and $\alpha$ is a hyper-parameter for controlling the ratio of this noise to the original reversed noise from the last video clip. The results of PNS are identical to those of FNR when $j<M$ and the difference is only on the remaining $N-M$ frames. New random noise sampled from a standard normal distribution is used when $\alpha=0$. A larger $\alpha$ brings more proportion of the reversed noise of the last video clip so more temporal consistency can be achieved.

\paragraph{Denoising with Staged Guidance.}

Though initial noises $\mathbf{z}_T$ are essential to the final denoised results as previous works~\citep{blattmann2023align,luo2023videofusion} have revealed, we still cannot get exactly the same video frames twice by independent denoising with DDIM~\citep{song2021denoising} sampling, even with the same initial noises. To improve the temporal consistency across continuous frames, e.g., between $\mathbf{z}_0^{i,N-1}$ and $\mathbf{z}_0^{i+1,0}$, reusing the initial noises created by FNR is not enough so Denoising with Staged Guidance (DSG) is proposed. Given an already generated $i$-th video clip, the purpose of DSG is to generate the starting video frames of $i+1$-th clip with high temporal consistency following the $M$ prompt frames of clip $i$. The approach of DSG involves replicating some denoising steps by reusing the latent features from the DDIM process in generating $i$-th clip. At the same time, to avoid the duplication of video frames but in reversed order, a staged strategy for denoising with guidance is proposed. Specifically, we have:
\begin{equation}
    \mathbf{z}_{t-1}^{i,j} = 
    \begin{cases}
        \mathbf{z}_{t-1}^{i-1, N-j-1} & \text{if } t > (1-\beta)T+\frac{\beta T j }{M} \\
        \text{DDIM}(\mathbf{z}_t^{i,j}, t) & \text{otherwise}
    \end{cases}, \;\;
    \beta \in [0, 1]
\end{equation}
where $\beta$ represents the extent of guided denoising. Latent features in the current clip are denoised totally with $\text{DDIM}$ sampling without any reference from the previous video clip when $\beta=0$, and a larger $\beta$ brings more guidance for each latent feature with $j < M$. In this way, video content of the first $M$ prompt frames can be consistent with the last $M$ frames of the last clip, and new content can also emerge along with the denoising process because staged guidance becomes weaker as $j$ gets closer to $M$.

\subsection{Temporal-aware decoder fine-tuning}
\label{subsec:vae_ft}

\begin{figure}[ht]
    \centering
    \includegraphics[width=.9\linewidth]{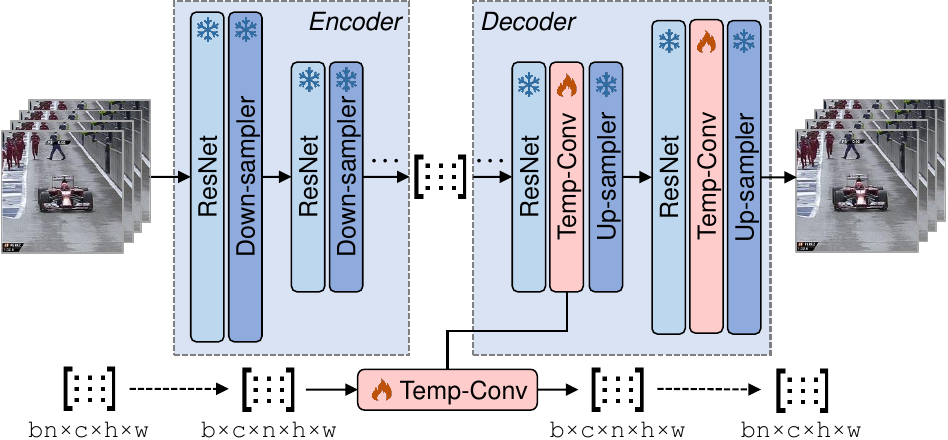}
    \caption{The autoencoder for latent representation is inherited and initialized from Stable Diffusion. \textit{Temp-Conv} representing temporal layers and implemented with 3D convolutions are injected into the decoder. Most parts of the autoencoder are frozen and only the parameters of \textit{Temp-Conv} are trainable.}
    \label{fig:method_vae}
\end{figure}

Since the original autoencoder of Stable Diffusion is specifically designed for image synthesis, it is necessary to fine-tune it with video data for better performance of video synthesis. However, the latent features used as inputs to U-Net after encoding are critical for efficient training when the pre-trained weights of Stable Diffusion are loaded. The autoencoder's architecture including its encoder and decoder is shown in Figure~\ref{fig:method_vae}. The encoder remains unchanged and the weights are frozen during fine-tuning. Also, temporal relations across video frames need to be considered for better temporal consistency after decoding so we add \textit{Temp-Conv} layers after \textit{ResNet} of each block in the decoder. For efficient fine-tuning, only the newly added \textit{Temp-Conv} layers are trainable. In addition, for better adapting from the autoencoder for image, we initialize the last layer of \textit{Temp-Conv} with zero and apply a residual connection.

For fine-tuning the autoencoder, we use the same datasets used for training U-Net describe in Section~\ref{sec:data_compose}. The fine-tuning also follows the adversarial manner following Stable Diffusion~\citep{rombach2022high}. The total loss is as follows:
\begin{equation}
    \mathcal{L} = \alpha_\text{rec}\mathcal{L}_\text{rec}(\mathbf{x}, \mathcal{D}(\mathcal{E}(\mathbf{x})) + \alpha_\text{reg}  \mathcal{L}_\text{reg}(\mathbf{x};\mathcal{E},\mathcal{D}) + \alpha_\text{disc} \mathcal{L}_\text{disc}(\mathcal{D}(\mathcal{E}(\mathbf{x}))
\end{equation}
In addition to the main reconstruction loss $\mathcal{L}_\text{rec}$ and a regularizing loss $\mathcal{L}_\text{reg}$ for regularizing the latent representation, a discrimination loss $\mathcal{L}_\text{disc}$ is also used which is computed by a patch-based discriminator for differentiating the original videos from the reconstructed. These three losses are respectively weighted with $\alpha_\text{rec}$, $\alpha_\text{reg}$ and $\alpha_\text{disc}$.

\section{Experiments}
\label{sec:experiments}

\subsection{Experimental setups}

\paragraph{Model architecture and sampling.} To exploit the ability of image synthesis models, we use the pre-trained weights of Stable Diffusion v2.1 to initialize the spatial layers of our model. Both the VAE and the text encoder are frozen after they are initialized with pre-trained weights from Stable Diffusion. During model training, only the newly added temporal layers and transformer blocks of the spatial layers are trainable. Since our model is essentially an LDM, VAE of Stable Diffusion but with a fine-tuned decoder is used for latent representation. For LDM sampling, we use DDIM~\citep{song2021denoising} in all our experiments.

\paragraph{Datasets.} In general, four types of datasets are used for training VidRD as shown in Table~\ref{tab:datasets}. \textbf{1) Well-captioned video-text datasets:} WebVid-2M~\citep{bain2021frozen}, TGIF~\citep{li2016tgif}, VATEX~\citep{wang2019vatex} and Pexels~\footnote{\url{https://huggingface.co/datasets/Corran/pexelvideos}}. WebVid-2M contains a total of about 2.5 million subtitled videos but we only use those whose duration is less than 20 seconds. Additionally, we use a basic watermark removal solution to remove watermarks from the videos. TGIF consists of 100K GIFs collected from Tumblr and the duration is relatively short. VATEX is a large-scale well-captioned video datasets covering 600 fine-grained human activities. Pexels contains about \num{360000} well-captioned videos from a popular website providing free stock videos. \textbf{2) Short video classification datasets:}  Moments-In-Time~\citep{monfort2021multi} and Kinetics-700~\citep{smaira2020short}. Moments-In-Time contains more than one million videos covering 339 action categories and Kinetics-700 contains over \num{650000} videos covering 700 human action categories. Most videos in these two datasets last a few seconds and each video is captioned using the strategy proposed in Section~\ref{sec:data_compose}. \textbf{3) Long video classification datasets:} VideoLT~\citep{zhang2021videolt}. This dataset contains a total of \num{250000} untrimmed long videos covering 1004 categories. After applying the strategy for long video classification datasets, we totally produce 800K captioned videos with an average length of 5 seconds. \textbf{4) Image datasets:} LAION-5B. This dataset originally contains 5.58 billion image-text pairs but only a small part of it is used as compensation to our video-text data. After applying the strategy of transforming images to videos introduced in Section~\ref{sec:data_compose}, 640K pseudo-videos with an average length of 2 seconds are produced.

\begin{table}[ht]
  \setlength{\tabcolsep}{1pt}
    \small
    \centering
    \begin{tabular}{l|cccc|cc|c|c}
       \toprule
       Dataset & WebVid-2M & TGIF & VATEX & Pexels & Moments-In-Time & Kinetics-700 & VideoLT & LAION-5B \\
       \midrule
        Num. Videos (K) & \num{1700} & \num{100} & \num{35} & \num{360} & \num{1000} & \num{650} & \num{800} & \num{640} \\
        Avg. Duration (s) & 11.9 & 3.1 & 10.0 & 19.5 & 3.0 & 10.0 & 5.0 & 2.0 \\
       \bottomrule
    \end{tabular}
    \caption{Four types of datasets are used including well-captioned video datasets, short video classification datasets, long video classification datasets, and image-text datasets.}
    \label{tab:datasets}
\end{table}

\paragraph{Training details.} VAE for encoding and decoding videos is the same as Stable Diffusion and only the newly added temporal layers in the decoder of VAE are trainable. For training the decoder of VAE, we set $\alpha_\text{rec}$, $\alpha_\text{reg}$ and $\alpha_\text{disc}$ to $1$, $1^{-5}$ and $0.5$ respectively. In U-Net, there are a total of 2.0B parameters, of which 565M are trainable and 316M are allocated for temporal layers. The base resolution of input videos for model training is $256\times256$ and 8 keyframes are sampled uniformly every 4 frames. The starting frame of sampling is randomly selected along the timeline. Each frame is resized along the shorter side and then randomly cropped to the target resolution. For video datasets with multiple captions such as VATEX~\citep{wang2019vatex}, one caption is randomly chosen every time it is sampled. In general, these practices can be seen as data augmentation.

\paragraph{Evaluation metrics.} Following previous works like Make-A-Video~\citep{singer2023makeavideo}, PYoCo~\citep{ge2023pyoco} and Video LDM~\citep{blattmann2023align}, the following metrics for quantitative evaluation are used: \textit{(i)} Fr\'echet Video Distance (FVD)~\citep{unterthiner2019fvd}: Following Make-A-Video~\citep{singer2023makeavideo}, we use a trained I3D model~\citep{carreira2017quo} for calculating FVD. \textit{(ii)} Inception Score (IS)~\citep{saito2020train}: Following previous works~\citep{singer2023makeavideo,hong2023cogvideo,blattmann2023align}, a trained C3D model~\citep{tran2015learning} is used for calculating the video version of IS.

\subsection{Main results}

To fully evaluate VidRD, we conduct both quantitative and qualitative evaluations. All the generated videos for evaluation are 16 frames in $256\times256$ resolution unless otherwise specified.

\begin{table}[ht]
    \setlength{\tabcolsep}{12pt}
    \centering
    \begin{tabular}{lrrr}
       \toprule
       Model & \#Videos for Training & IS $\uparrow$ & FVD $\downarrow$ \\
       \midrule
       CogVideo~\citep{hong2023cogvideo} & 5.4M & 25.27 & 701.59 \\
       MagicVideo~\citep{zhou2022magicvideo} & 27.0M & - & 699.00 \\
       LVDM~\citep{he2022latent} & 2.0M & - & 641.80 \\
       ModelScope~\citep{wang2023modelscope} & - & - & 639.90 \\
       Video LDM~\citep{blattmann2023align} & 10.7M & 33.45 & 550.61 \\
       Make-A-Video~\citep{singer2023makeavideo} & 20.0M & 33.00 & 367.23 \\
       VideoFactory~\citep{wang2023videofactory} & 140.7M & - & 410.00 \\
       \midrule
       VidRD w/o fine-tuned VAE & 5.3M & 39.24 & 369.48 \\
       VidRD w/ fine-tuned VAE & 5.3M & \textbf{39.37} & \textbf{363.19} \\
       \bottomrule
    \end{tabular}
    \caption{Quantitative evaluation results on UCF-101. All the videos for evaluation are generated in a zero-shot manner. In comparison with other methods, VidRD achieves better IS and FVD while using fewer videos for model training.}
    \label{tab:eval_ucf101}
\end{table}

\paragraph{Quantitative Evaluation.} Following previous works~\citep{singer2023makeavideo, hong2023cogvideo, blattmann2023align}, we use UCF-101~\citep{soomro2012ucf101}, a dataset for video recognition, for evaluating FVD and IS. Since there are only 101 brief class names such as \textit{knitting} and \textit{diving} in UCF-101, we devise a descriptive prompt for each class for video synthesis in our experiments. The whole list of prompts we use is provided in the supplementary material. Following Make-A-Video~\citep{singer2023makeavideo}, 10K videos are generated by VidRD following the same class distribution as the training set. It is worth noting that the experimental settings in VideoFactory~\citep{wang2023videofactory} are slightly different from VidRD. VideoFactory generates 100 samples for each class. The quantitative evaluation results are shown in Table~\ref{tab:eval_ucf101}. VidRD achieves the best FVD and IS while using much fewer videos for training. Meanwhile, fine-tuning the decoder of VAE helps improve VidRD further. The reason is that a temporal-aware decoder can make restoring pixels from latent features more accurate.

\begin{figure}[ht]
    \centering
    \includegraphics[width=.95\linewidth]{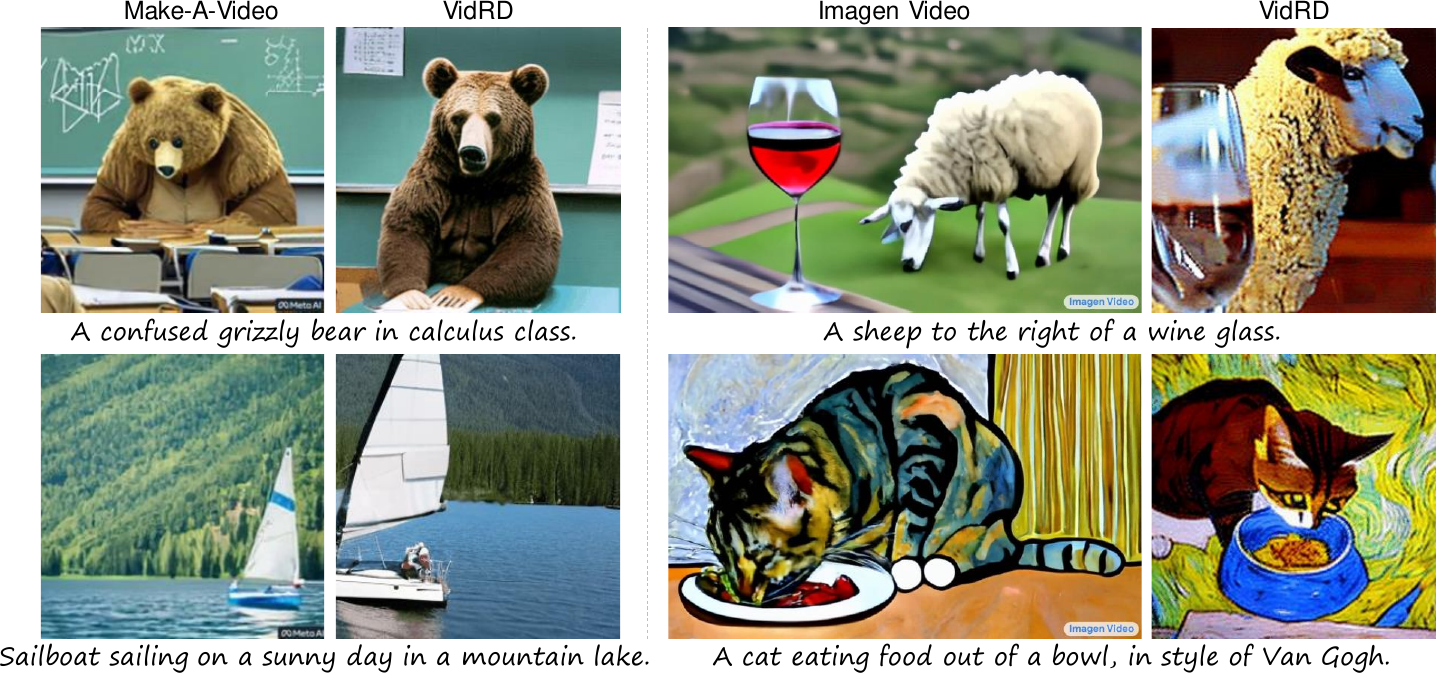}
    \caption{For comparison, some video examples generated by different methods are shown here. The examples generated by VidRD show good text alignment and structure.}
    \label{fig:eval_qualitative}
\end{figure}

\paragraph{Qualitative Evaluation.} Since all the currently used metrics for evaluating video generation models are considered not fully reliable and may be inconsistent with perception, qualitative evaluation is necessary. To this end, example videos are generated with the same text prompts by VidRD and the other models including Make-A-Video~\citep{singer2023makeavideo}, Imagen Video~\citep{ho2022imagenvideo}. Figure~\ref{fig:eval_qualitative} shows the comparisons between the video generation results of these methods. VidRD performs well in both structure and appearance. More video examples can be found on our project website.

\subsection{Ablation study}

\begin{figure}[ht]
  \centering
  \begin{subfigure}[t]{.42\linewidth}
    \includegraphics[width=\linewidth]{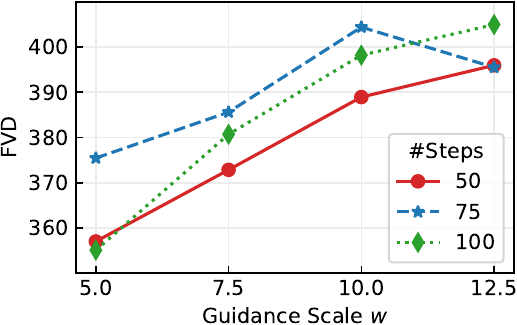}
    \caption{FVD of using different guidance scale $w$ and the number of inference steps.}
  \end{subfigure}
  \hspace{0.1\linewidth} 
  \begin{subfigure}[t]{.42\linewidth}
    \includegraphics[width=\linewidth]{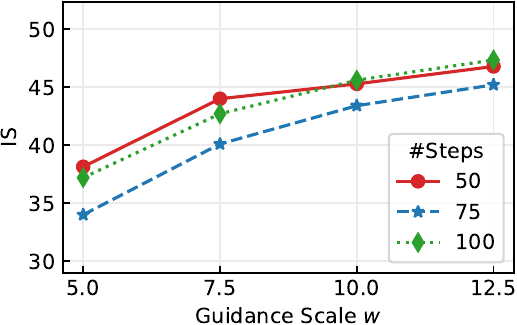}
    \caption{IS of using different guidance scales $w$ and the number of inference steps.}
  \end{subfigure}
  \caption{Ablation results of the number of inference steps and the guidance scale are based on FVD and IS, which are evaluated on UCF-101 in a zero-shot manner.}
  \label{fig:exp_guidance}
\end{figure}

\paragraph{Classifier-free guidance scale.} For generating diverse videos, two hyper-parameters are critical in model inference: the number of inference steps and the scale of classifier-free guidance. The number of inference steps means the total steps of denoising from the initial random noises to the resulting latent features of video frames. The scale of classifier-free guidance is proposed with classifier-free diffusion guidance~\citep{ho2019classifier}. In each reversed diffusion step during model inference, the predicted noise $\tilde{\mathbf{\epsilon}}_\theta$ is computed with two types of predictions: the prediction conditioned on prompt text features $\mathbf{c}$ that is $\mathbf{\epsilon}_\theta(\mathbf{z}, \mathbf{c})$ and the prediction without such condition that is $\mathbf{\epsilon}_\theta(\mathbf{z})$. The final predicted noise is calculated by combining these two, controlled with guidance scale $w$: $\tilde{\mathbf{\epsilon}}_\theta = \mathbf{\epsilon}_\theta(\mathbf{z}) + w (\mathbf{\epsilon}_\theta(\mathbf{z}, \mathbf{c}) - \mathbf{\epsilon}_\theta(\mathbf{z}))$. Classifier-free guidance is disabled when $w=1$ and a larger $w$ means more video-text alignment but weaker diversity. To study the effects of these two hyper-parameters only, we do experiments using our base model without any strategies for iterative video generation. Figure~\ref{fig:exp_guidance} shows FVD and IS results of using different combinations of these hyper-parameters. The impact of the number of inference steps is relatively small and we set the number to 50 considering the efficiency of video synthesis. For the guidance scale, there is a trade-off between FVD and IS. A smaller guidance scale can achieve lower FVD which means higher temporal consistency. Yet a small guidance scale leads to low IS which means poor spatial appearance. Therefore, we set the guidance scale to $w=10.0$.

\paragraph{Joint training with image datasets.} To evaluate the effect of VidRD of joint training with pseudo-videos produced from image-text datasets, experiments are designed by fine-tuning VidRD with a small-scale dataset in which images are either in raw format or in pseudo-video style. This dataset totally consists of \num{5000} videos from the VATEX~\citep{wang2019vatex} testing set and \num{8000} images from LAION-5B~\citep{schuhmann2022laion}. We tabulate our findings in Table~\ref{tab:eval_images}. Compared with using static images for training spatial layers only, we find that pseudo-videos, produced by random zooming and panning of static images, help enhance temporal consistency but compromise visual appearance. The reason is that pseudo-videos bring more training for temporal layers while static images focus on spatial layers more. In practice, since the amount of video data is much less than that of image data, this technique is a cheap way to train a video diffusion model with high temporal consistency.

\begin{table}[t]
  \setlength{\tabcolsep}{15pt}
    \centering
    \begin{tabular}{lcc}
       \toprule
       Strategy  & IS $\uparrow$ & FVD $\downarrow$ \\
       \midrule
        VidRD w/o pseudo-videos & \textbf{42.00} & 451.16 \\
        VidRD w/ pseudo-videos & 40.87 & \textbf{433.22} \\
       \bottomrule
    \end{tabular}
    \caption{This is a comparison between fine-tuning VidRD with and without pseudo-videos by random zooming and panning static images. FVD and IS are evaluated on UCF-101 and all experiments here are in a zero-shot manner.}
    \label{tab:eval_images}
\end{table}

\begin{figure}[ht]
  \centering
  \begin{subfigure}[t]{.32\linewidth}
    \includegraphics[width=\linewidth]{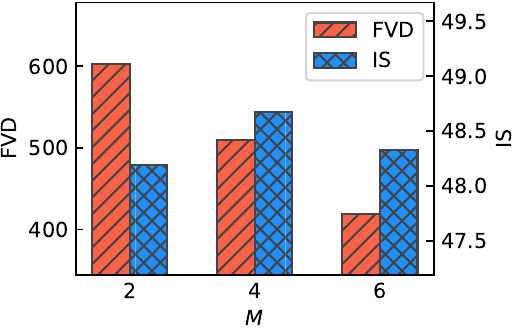}
    \caption{Ablation studies on $M$ with $\alpha=4$ and $\beta=0.4$.}
    \label{fig:exp_abl_m}
  \end{subfigure}
  \begin{subfigure}[t]{.32\linewidth}
    \includegraphics[width=\linewidth]{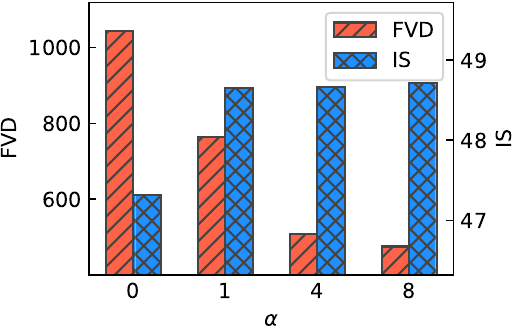}
    \caption{Ablation studies on $\alpha$ in PNS with $M=4$ and $\beta=0.4$.}
    \label{fig:exp_abl_alpha}
  \end{subfigure}
  \begin{subfigure}[t]{.32\linewidth}
    \includegraphics[width=\linewidth]{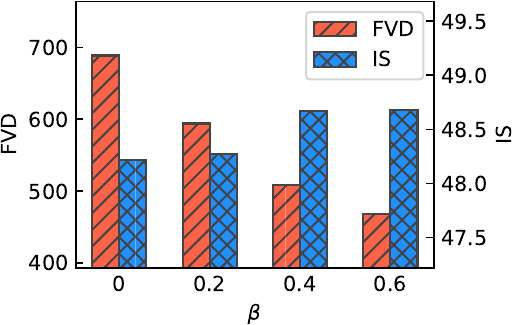}
    \caption{Ablation studies on $\beta$ in DSG with $M=4$ and $\alpha=4$.}
    \label{fig:exp_abl_beta}
  \end{subfigure}
  \caption{Ablation studies on hyper-parameters for a text-to-video generation under guidance scale set to $10$ and the number of inference steps set to $50$.}
  \label{fig:exp_ablation}
\end{figure}

\paragraph{Hyper-parameters for inference.} Revisiting Algorithm~\ref{alg:iterative_generation}, there are three critical inference parameters influencing our iterative video generation: $M \in \{0, 1, \dots, N\}$ representing the number of prompt frames, $\alpha \geq 0$ controlling the degree to which the previous noise is reused and $\beta \in [0,1]$ controlling the proportion of denoising guidance. Following the controlled variable method, we conduct ablation studies on these three parameters. The ablation results are shown in Figure~\ref{fig:exp_ablation}. Figure~\ref{fig:exp_abl_m} shows the parameter $M$ has a significant impact on temporal consistency but has little effect on structural quality. While a larger value of parameter $M$ can help achieve better quantitative metrics, it also leads to more frequent occurrences of content cycling across video clips. The ablation result for $\alpha$ is shown in Figure~\ref{fig:exp_abl_alpha}. We find that temporal consistency can be improved by reusing more noises from the previous clip. Also, the appearance or structure quality reflected by IS can be improved once the noise is reused, that is, $\alpha>0$. For $\beta$ whose ablation result is shown in Figure~\ref{fig:exp_abl_beta}, both FVD and IS show improvement so DSG is necessary for generating smooth videos. However, in practice, we find that too large a $\beta$ is more likely to lead to content cycling across video clips, which can be measured by neither FVD nor IS. As a result, we set $\beta=0.4$ to get a balance between video smoothness and inception degradation caused by content cycling.

\begin{figure}[ht]
    \centering
    \includegraphics[width=\linewidth]{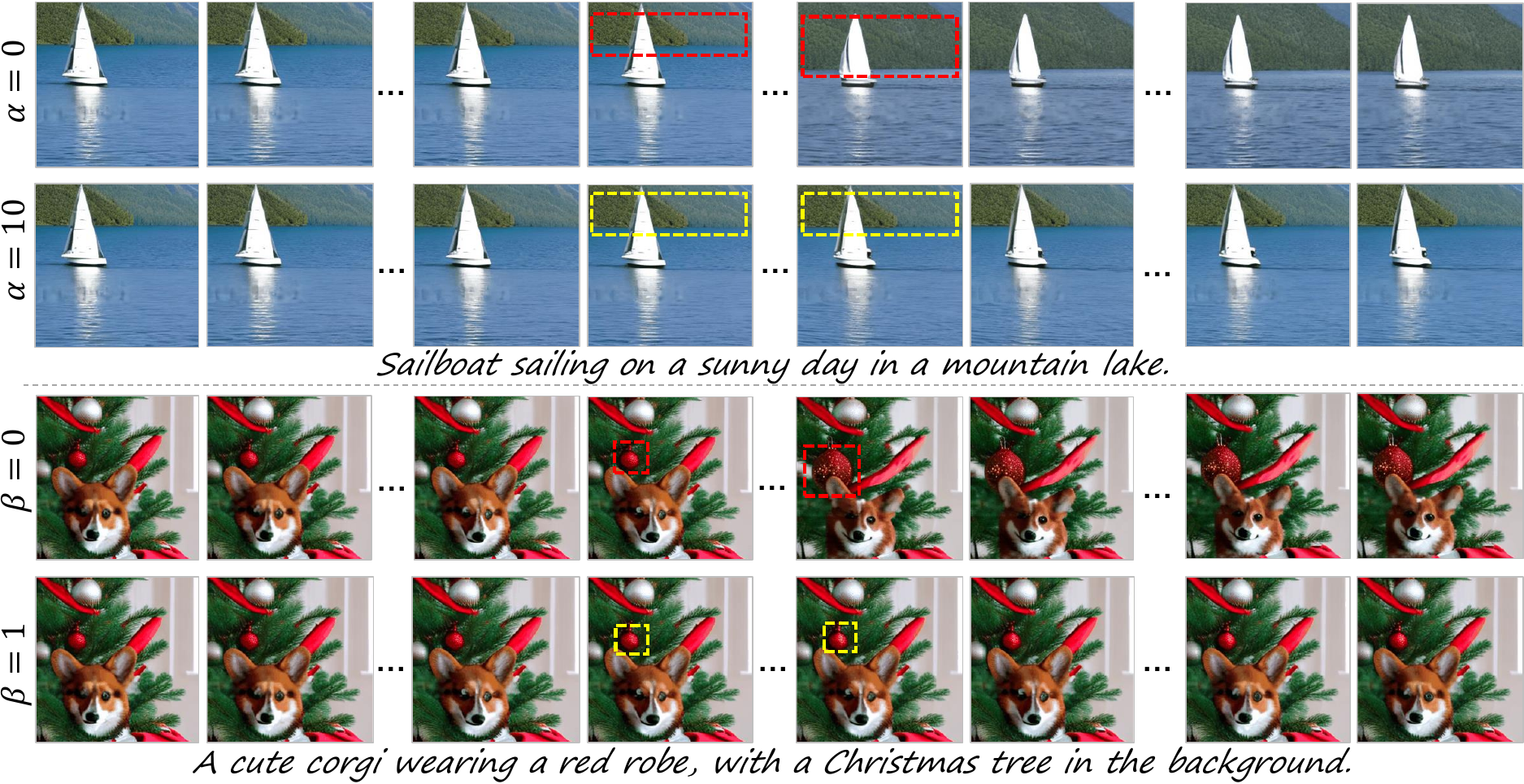}
    \caption{These are two visual examples showing the generated videos using different values of $\alpha$ and $\beta$ respectively. The red dashed box highlights the positions where abrupt, unreasonable visual changes occur. The yellow dashed box indicates that the issue is resolved by enabling PNS and DSG.}
    \label{fig:eval_abl_demos}
\end{figure}

To gain a more intuitive understanding of the parameters $\alpha$ and $\beta$, Figure~\ref{fig:eval_abl_demos} shows some generated videos with different parameter values. Some unreasonable abrupt video content changes, highlighted with red dashed boxes in the figure, can be easily observed when $\alpha=0$ in which case no guiding steps are present for denoising the prompt frames in each video clip. Similar artifacts can also be observed when $\beta=0$ in which case there are no guiding steps for denoising the prompt frames.

\section{Conclusion}
\label{sec:conclusion}



In this work, we introduce a novel text-to-video framework called VidRD to generate smooth videos with text guidance. A set of strategies is proposed to exploit multiple existing datasets, which include video datasets for action recognition and image-text datasets, in order to train our text-to-video generation model. For generating longer videos, we propose an iterative approach through reusing the noise. and imitating the diffusion process clip-by-clip. Additionally, the autoencoder is improved by injecting temporal layers to generate smooth videos with high temporal consistency. Extensive experiments demonstrate that our method excels in both quantitative and qualitative evaluations.



\bibliographystyle{iclr2024_conference}
\bibliography{ms}

\begin{thebibliography}{69}
\providecommand{\natexlab}[1]{#1}
\providecommand{\url}[1]{\texttt{#1}}
\expandafter\ifx\csname urlstyle\endcsname\relax
  \providecommand{\doi}[1]{doi: #1}\else
  \providecommand{\doi}{doi: \begingroup \urlstyle{rm}\Url}\fi

\bibitem[Acharya et~al.(2018)Acharya, Huang, Paudel, and
  Van~Gool]{acharya2018towards}
Dinesh Acharya, Zhiwu Huang, Danda~Pani Paudel, and Luc Van~Gool.
\newblock Towards high resolution video generation with progressive growing of
  sliced wasserstein gans.
\newblock \emph{arXiv preprint arXiv:1810.02419}, 2018.

\bibitem[Bain et~al.(2021)Bain, Nagrani, Varol, and Zisserman]{bain2021frozen}
Max Bain, Arsha Nagrani, G{\"u}l Varol, and Andrew Zisserman.
\newblock Frozen in time: A joint video and image encoder for end-to-end
  retrieval.
\newblock In \emph{Proceedings of the IEEE/CVF International Conference on
  Computer Vision ({ICCV})}, pp.\  1728--1738, 2021.

\bibitem[Bhunia et~al.(2023)Bhunia, Khan, Cholakkal, Anwer, Laaksonen, Shah,
  and Khan]{bhunia2023person}
Ankan~Kumar Bhunia, Salman Khan, Hisham Cholakkal, Rao~Muhammad Anwer, Jorma
  Laaksonen, Mubarak Shah, and Fahad~Shahbaz Khan.
\newblock Person image synthesis via denoising diffusion model.
\newblock In \emph{Proceedings of the IEEE/CVF Conference on Computer Vision
  and Pattern Recognition ({CVPR})}, pp.\  5968--5976, 2023.

\bibitem[Blattmann et~al.(2023)Blattmann, Rombach, Ling, Dockhorn, Kim, Fidler,
  and Kreis]{blattmann2023align}
Andreas Blattmann, Robin Rombach, Huan Ling, Tim Dockhorn, Seung~Wook Kim,
  Sanja Fidler, and Karsten Kreis.
\newblock Align your latents: High-resolution video synthesis with latent
  diffusion models.
\newblock In \emph{Proceedings of the IEEE/CVF Conference on Computer Vision
  and Pattern Recognition ({CVPR})}, pp.\  22563--22575, 2023.

\bibitem[Brooks et~al.(2022)Brooks, Hellsten, Aittala, Wang, Aila, Lehtinen,
  Liu, Efros, and Karras]{brooks2022generating}
Tim Brooks, Janne Hellsten, Miika Aittala, Ting-Chun Wang, Timo Aila, Jaakko
  Lehtinen, Ming-Yu Liu, Alexei Efros, and Tero Karras.
\newblock Generating long videos of dynamic scenes.
\newblock \emph{Advances in Neural Information Processing Systems ({NeurIPS})},
  35:\penalty0 31769--31781, 2022.

\bibitem[Carreira \& Zisserman(2017)Carreira and Zisserman]{carreira2017quo}
Joao Carreira and Andrew Zisserman.
\newblock Quo vadis, action recognition? a new model and the kinetics dataset.
\newblock In \emph{Proceedings of the IEEE Conference on Computer Vision and
  Pattern Recognition ({CVPR})}, pp.\  6299--6308, 2017.

\bibitem[Chang et~al.(2022)Chang, Zhang, Jiang, Liu, and
  Freeman]{chang2022maskgit}
Huiwen Chang, Han Zhang, Lu~Jiang, Ce~Liu, and William~T Freeman.
\newblock Maskgit: Masked generative image transformer.
\newblock In \emph{Proceedings of the IEEE/CVF Conference on Computer Vision
  and Pattern Recognition ({CVPR})}, pp.\  11315--11325, 2022.

\bibitem[Clune(2019)]{clune2019aigas}
Jeff Clune.
\newblock Ai-gas: Ai-generating algorithms, an alternate paradigm for producing
  general artificial intelligence.
\newblock \emph{arXiv preprint arXiv:1905.10985}, 2019.

\bibitem[Dabral et~al.(2023)Dabral, Mughal, Golyanik, and
  Theobalt]{dabral2023mofusion}
Rishabh Dabral, Muhammad~Hamza Mughal, Vladislav Golyanik, and Christian
  Theobalt.
\newblock Mofusion: A framework for denoising-diffusion-based motion synthesis.
\newblock In \emph{Proceedings of the IEEE/CVF Conference on Computer Vision
  and Pattern Recognition ({CVPR})}, pp.\  9760--9770, 2023.

\bibitem[Ding et~al.(2021)Ding, Yang, Hong, Zheng, Zhou, Yin, Lin, Zou, Shao,
  Yang, et~al.]{ding2021cogview}
Ming Ding, Zhuoyi Yang, Wenyi Hong, Wendi Zheng, Chang Zhou, Da~Yin, Junyang
  Lin, Xu~Zou, Zhou Shao, Hongxia Yang, et~al.
\newblock Cogview: Mastering text-to-image generation via transformers.
\newblock \emph{Advances in Neural Information Processing Systems ({NeurIPS})},
  34:\penalty0 19822--19835, 2021.

\bibitem[Esser et~al.(2023)Esser, Chiu, Atighehchian, Granskog, and
  Germanidis]{esser2023structure}
Patrick Esser, Johnathan Chiu, Parmida Atighehchian, Jonathan Granskog, and
  Anastasis Germanidis.
\newblock Structure and content-guided video synthesis with diffusion models.
\newblock \emph{arXiv preprint arXiv:2302.03011}, 2023.

\bibitem[Fan et~al.(2023)Fan, Chen, Chen, Cheng, Yuan, and Wang]{fan2023frido}
Wan-Cyuan Fan, Yen-Chun Chen, DongDong Chen, Yu~Cheng, Lu~Yuan, and
  Yu-Chiang~Frank Wang.
\newblock Frido: Feature pyramid diffusion for complex scene image synthesis.
\newblock In \emph{Proceedings of the AAAI Conference on Artificial
  Intelligence}, volume~37, pp.\  579--587, 2023.

\bibitem[Fjelland(2020)]{fjelland2020general}
Ragnar Fjelland.
\newblock Why general artificial intelligence will not be realized.
\newblock \emph{Humanities and Social Sciences Communications}, 7\penalty0
  (1):\penalty0 1--9, 2020.

\bibitem[Ge et~al.(2022)Ge, Hayes, Yang, Yin, Pang, Jacobs, Huang, and
  Parikh]{ge2022long}
Songwei Ge, Thomas Hayes, Harry Yang, Xi~Yin, Guan Pang, David Jacobs, Jia-Bin
  Huang, and Devi Parikh.
\newblock Long video generation with time-agnostic vqgan and time-sensitive
  transformer.
\newblock In \emph{European Conference on Computer Vision ({ECCV})}, pp.\
  102--118. Springer, 2022.

\bibitem[Ge et~al.(2023)Ge, Nah, Liu, Poon, Tao, Catanzaro, Jacobs, Huang, Liu,
  and Balaji]{ge2023pyoco}
Songwei Ge, Seungjun Nah, Guilin Liu, Tyler Poon, Andrew Tao, Bryan Catanzaro,
  David Jacobs, Jia-Bin Huang, Ming-Yu Liu, and Yogesh Balaji.
\newblock Preserve your own correlation: A noise prior for video diffusion
  models.
\newblock \emph{arXiv preprint arXiv:2305.10474}, 2023.

\bibitem[Goertzel \& Pennachin(2007)Goertzel and
  Pennachin]{goertzel2007artificial}
Ben Goertzel and Cassio Pennachin.
\newblock \emph{Artificial general intelligence}, volume~2.
\newblock Springer, 2007.

\bibitem[Harvey et~al.(2022)Harvey, Naderiparizi, Masrani, Weilbach, and
  Wood]{harvey2022flexible}
William Harvey, Saeid Naderiparizi, Vaden Masrani, Christian Weilbach, and
  Frank Wood.
\newblock Flexible diffusion modeling of long videos.
\newblock \emph{Advances in Neural Information Processing Systems ({NeurIPS})},
  35:\penalty0 27953--27965, 2022.

\bibitem[He et~al.(2022)He, Yang, Zhang, Shan, and Chen]{he2022latent}
Yingqing He, Tianyu Yang, Yong Zhang, Ying Shan, and Qifeng Chen.
\newblock Latent video diffusion models for high-fidelity video generation with
  arbitrary lengths.
\newblock \emph{arXiv preprint arXiv:2211.13221}, 2022.

\bibitem[Ho \& Salimans(2021)Ho and Salimans]{ho2019classifier}
Jonathan Ho and Tim Salimans.
\newblock Classifier-free diffusion guidance.
\newblock In \emph{NeurIPS 2021 Workshop on Deep Generative Models and
  Downstream Applications}, 2021.

\bibitem[Ho et~al.(2022)Ho, Chan, Saharia, Whang, Gao, Gritsenko, Kingma,
  Poole, Norouzi, Fleet, et~al.]{ho2022imagenvideo}
Jonathan Ho, William Chan, Chitwan Saharia, Jay Whang, Ruiqi Gao, Alexey
  Gritsenko, Diederik~P Kingma, Ben Poole, Mohammad Norouzi, David~J Fleet,
  et~al.
\newblock Imagen video: High definition video generation with diffusion models.
\newblock \emph{arXiv preprint arXiv:2210.02303}, 2022.

\bibitem[Hong et~al.(2023)Hong, Ding, Zheng, Liu, and Tang]{hong2023cogvideo}
Wenyi Hong, Ming Ding, Wendi Zheng, Xinghan Liu, and Jie Tang.
\newblock Cogvideo: Large-scale pretraining for text-to-video generation via
  transformers.
\newblock In \emph{International Conference on Learning Representations
  ({ICLR})}, 2023.

\bibitem[Huang et~al.(2023)Huang, Chen, Liu, Shen, Zhao, and
  Zhou]{huang2023composer}
Lianghua Huang, Di~Chen, Yu~Liu, Yujun Shen, Deli Zhao, and Jingren Zhou.
\newblock Composer: Creative and controllable image synthesis with composable
  conditions.
\newblock \emph{arXiv preprint arXiv:2302.09778}, 2023.

\bibitem[Karras et~al.(2019)Karras, Laine, and Aila]{karras2019style}
Tero Karras, Samuli Laine, and Timo Aila.
\newblock A style-based generator architecture for generative adversarial
  networks.
\newblock In \emph{Proceedings of the IEEE/CVF Conference on Computer Vision
  and Pattern Recognition ({CVPR})}, pp.\  4401--4410, 2019.

\bibitem[Kawar et~al.(2023)Kawar, Ganz, and Elad]{kawar2023enhancing}
Bahjat Kawar, Roy Ganz, and Michael Elad.
\newblock Enhancing diffusion-based image synthesis with robust classifier
  guidance.
\newblock \emph{Transactions on Machine Learning Research}, 2023.
\newblock ISSN 2835-8856.

\bibitem[Khachatryan et~al.(2023)Khachatryan, Movsisyan, Tadevosyan, Henschel,
  Wang, Navasardyan, and Shi]{khachatryan2023text2video}
Levon Khachatryan, Andranik Movsisyan, Vahram Tadevosyan, Roberto Henschel,
  Zhangyang Wang, Shant Navasardyan, and Humphrey Shi.
\newblock Text2video-zero: Text-to-image diffusion models are zero-shot video
  generators.
\newblock \emph{arXiv preprint arXiv:2303.13439}, 2023.

\bibitem[Le~Moing et~al.(2021)Le~Moing, Ponce, and Schmid]{moing2021ccvs}
Guillaume Le~Moing, Jean Ponce, and Cordelia Schmid.
\newblock Ccvs: context-aware controllable video synthesis.
\newblock \emph{Advances in Neural Information Processing Systems ({NeurIPS})},
  34:\penalty0 14042--14055, 2021.

\bibitem[Li et~al.(2019)Li, Qi, Lukasiewicz, and Torr]{li2019controllable}
Bowen Li, Xiaojuan Qi, Thomas Lukasiewicz, and Philip Torr.
\newblock Controllable text-to-image generation.
\newblock \emph{Advances in Neural Information Processing Systems ({NeurIPS})},
  32, 2019.

\bibitem[Li et~al.(2023)Li, Li, Savarese, and Hoi]{li2023blip}
Junnan Li, Dongxu Li, Silvio Savarese, and Steven Hoi.
\newblock Blip-2: Bootstrapping language-image pre-training with frozen image
  encoders and large language models.
\newblock \emph{arXiv preprint arXiv:2301.12597}, 2023.

\bibitem[Li et~al.(2016)Li, Song, Cao, Tetreault, Goldberg, Jaimes, and
  Luo]{li2016tgif}
Yuncheng Li, Yale Song, Liangliang Cao, Joel Tetreault, Larry Goldberg,
  Alejandro Jaimes, and Jiebo Luo.
\newblock {TGIF: A New Dataset and Benchmark on Animated GIF Description}.
\newblock In \emph{The IEEE Conference on Computer Vision and Pattern
  Recognition (CVPR)}, June 2016.

\bibitem[Liu et~al.(2023{\natexlab{a}})Liu, Zhang, Li, Lin, and
  Jia]{liu2023videop2p}
Shaoteng Liu, Yuechen Zhang, Wenbo Li, Zhe Lin, and Jiaya Jia.
\newblock Video-p2p: Video editing with cross-attention control.
\newblock \emph{arXiv preprint arXiv:2303.04761}, 2023{\natexlab{a}}.

\bibitem[Liu et~al.(2023{\natexlab{b}})Liu, Park, Azadi, Zhang, Chopikyan, Hu,
  Shi, Rohrbach, and Darrell]{liu2023more}
Xihui Liu, Dong~Huk Park, Samaneh Azadi, Gong Zhang, Arman Chopikyan, Yuxiao
  Hu, Humphrey Shi, Anna Rohrbach, and Trevor Darrell.
\newblock More control for free! image synthesis with semantic diffusion
  guidance.
\newblock In \emph{Proceedings of the IEEE/CVF Winter Conference on
  Applications of Computer Vision ({CACV})}, pp.\  289--299,
  2023{\natexlab{b}}.

\bibitem[Luo et~al.(2023)Luo, Chen, Zhang, Huang, Wang, Shen, Zhao, Zhou, and
  Tan]{luo2023videofusion}
Zhengxiong Luo, Dayou Chen, Yingya Zhang, Yan Huang, Liang Wang, Yujun Shen,
  Deli Zhao, Jingren Zhou, and Tieniu Tan.
\newblock Videofusion: Decomposed diffusion models for high-quality video
  generation.
\newblock In \emph{Proceedings of the IEEE/CVF Conference on Computer Vision
  and Pattern Recognition ({CVPR})}, pp.\  10209--10218, 2023.

\bibitem[Molad et~al.(2023)Molad, Horwitz, Valevski, Acha, Matias, Pritch,
  Leviathan, and Hoshen]{molad2023dreamix}
Eyal Molad, Eliahu Horwitz, Dani Valevski, Alex~Rav Acha, Yossi Matias, Yael
  Pritch, Yaniv Leviathan, and Yedid Hoshen.
\newblock Dreamix: Video diffusion models are general video editors.
\newblock \emph{arXiv preprint arXiv:2302.01329}, 2023.

\bibitem[Monfort et~al.(2021)Monfort, Pan, Ramakrishnan, Andonian, McNamara,
  Lascelles, Fan, Gutfreund, Feris, and Oliva]{monfort2021multi}
Mathew Monfort, Bowen Pan, Kandan Ramakrishnan, Alex Andonian, Barry~A
  McNamara, Alex Lascelles, Quanfu Fan, Dan Gutfreund, Rog{\'e}rio~Schmidt
  Feris, and Aude Oliva.
\newblock Multi-moments in time: Learning and interpreting models for
  multi-action video understanding.
\newblock \emph{IEEE Transactions on Pattern Analysis and Machine
  Intelligence}, 44\penalty0 (12):\penalty0 9434--9445, 2021.

\bibitem[Nichol et~al.(2021)Nichol, Dhariwal, Ramesh, Shyam, Mishkin, McGrew,
  Sutskever, and Chen]{nichol2021glide}
Alex Nichol, Prafulla Dhariwal, Aditya Ramesh, Pranav Shyam, Pamela Mishkin,
  Bob McGrew, Ilya Sutskever, and Mark Chen.
\newblock {GLIDE}: Towards photorealistic image generation and editing with
  text-guided diffusion models.
\newblock \emph{arXiv preprint arXiv:2112.10741}, 2021.

\bibitem[Qi et~al.(2023)Qi, Cun, Zhang, Lei, Wang, Shan, and
  Chen]{qi2023fatezero}
Chenyang Qi, Xiaodong Cun, Yong Zhang, Chenyang Lei, Xintao Wang, Ying Shan,
  and Qifeng Chen.
\newblock Fatezero: Fusing attentions for zero-shot text-based video editing.
\newblock \emph{arXiv preprint arXiv:2303.09535}, 2023.

\bibitem[Radford et~al.(2021)Radford, Kim, Hallacy, Ramesh, Goh, Agarwal,
  Sastry, Askell, Mishkin, Clark, et~al.]{radford2021learning}
Alec Radford, Jong~Wook Kim, Chris Hallacy, Aditya Ramesh, Gabriel Goh,
  Sandhini Agarwal, Girish Sastry, Amanda Askell, Pamela Mishkin, Jack Clark,
  et~al.
\newblock Learning transferable visual models from natural language
  supervision.
\newblock In \emph{International Conference on Machine Learning ({ICML})}, pp.\
   8748--8763. PMLR, 2021.

\bibitem[Ramesh et~al.(2021)Ramesh, Pavlov, Goh, Gray, Voss, Radford, Chen, and
  Sutskever]{ramesh2021zeroshot}
Aditya Ramesh, Mikhail Pavlov, Gabriel Goh, Scott Gray, Chelsea Voss, Alec
  Radford, Mark Chen, and Ilya Sutskever.
\newblock Zero-shot text-to-image generation.
\newblock In \emph{International Conference on Machine Learning ({ICML})}, pp.\
   8821--8831. PMLR, 2021.

\bibitem[Ramesh et~al.(2022)Ramesh, Dhariwal, Nichol, Chu, and
  Chen]{ramech2022hierarchical}
Aditya Ramesh, Prafulla Dhariwal, Alex Nichol, Casey Chu, and Mark Chen.
\newblock Hierarchical text-conditional image generation with clip latents.
\newblock \emph{arXiv preprint arXiv:2204.06125}, 2022.

\bibitem[Rombach et~al.(2022)Rombach, Blattmann, Lorenz, Esser, and
  Ommer]{rombach2022high}
Robin Rombach, Andreas Blattmann, Dominik Lorenz, Patrick Esser, and Bj{\"o}rn
  Ommer.
\newblock High-resolution image synthesis with latent diffusion models.
\newblock In \emph{Proceedings of the IEEE/CVF Conference on Computer Vision
  and Pattern Recognition ({CVPR})}, pp.\  10684--10695, 2022.

\bibitem[Saharia et~al.(2022)Saharia, Chan, Saxena, Li, Whang, Denton, Kamyar,
  Ghasemipour, Karagol, Mahdavi, Lopes, Salimans, Ho, Fleet, and
  Norouzi]{saharia2022photorealistic}
Chitwan Saharia, William Chan, Saurabh Saxena, Lala Li, Jay Whang, Emily
  Denton, Seyed Kamyar, Seyed Ghasemipour, Burcu Karagol, SSara Mahdavi,
  RaphaGontijo Lopes, Tim Salimans, Jonathan Ho, DavidJ Fleet, and Mohammad
  Norouzi.
\newblock Photorealistic text-to-image diffusion models with deep language
  understanding.
\newblock \emph{Advances in Neural Information Processing Systems ({NeurIPS})},
  35:\penalty0 36479--36494, 2022.

\bibitem[Saito et~al.(2017)Saito, Matsumoto, and Saito]{saito2017temporal}
Masaki Saito, Eiichi Matsumoto, and Shunta Saito.
\newblock Temporal generative adversarial nets with singular value clipping.
\newblock In \emph{Proceedings of the IEEE International Conference on Computer
  Vision ({ICCV})}, pp.\  2830--2839, 2017.

\bibitem[Saito et~al.(2020)Saito, Saito, Koyama, and Kobayashi]{saito2020train}
Masaki Saito, Shunta Saito, Masanori Koyama, and Sosuke Kobayashi.
\newblock Train sparsely, generate densely: Memory-efficient unsupervised
  training of high-resolution temporal {GAN}.
\newblock \emph{International Journal of Computer Vision}, 128\penalty0
  (10):\penalty0 2586--2606, 2020.

\bibitem[Schuhmann et~al.(2022)Schuhmann, Beaumont, Vencu, Gordon, Wightman,
  Cherti, Coombes, Katta, Mullis, Wortsman, et~al.]{schuhmann2022laion}
Christoph Schuhmann, Romain Beaumont, Richard Vencu, Cade Gordon, Ross
  Wightman, Mehdi Cherti, Theo Coombes, Aarush Katta, Clayton Mullis, Mitchell
  Wortsman, et~al.
\newblock Laion-5b: An open large-scale dataset for training next generation
  image-text models.
\newblock \emph{Advances in Neural Information Processing Systems ({NeurIPS})},
  35:\penalty0 25278--25294, 2022.

\bibitem[Singer et~al.(2023)Singer, Polyak, Hayes, Yin, An, Zhang, Hu, Yang,
  Ashual, Gafni, Parikh, Gupta, and Taigman]{singer2023makeavideo}
Uriel Singer, Adam Polyak, Thomas Hayes, Xi~Yin, Jie An, Songyang Zhang, Qiyuan
  Hu, Harry Yang, Oron Ashual, Oran Gafni, Devi Parikh, Sonal Gupta, and Yaniv
  Taigman.
\newblock Make-a-video: Text-to-video generation without text-video data.
\newblock In \emph{International Conference on Learning Representations
  ({ICLR})}, 2023.

\bibitem[Skorokhodov et~al.(2022)Skorokhodov, Tulyakov, and
  Elhoseiny]{skorokhodov2022styleganv}
Ivan Skorokhodov, Sergey Tulyakov, and Mohamed Elhoseiny.
\newblock {StyleGAN-V}: A continuous video generator with the price, image
  quality and perks of stylegan2.
\newblock In \emph{Proceedings of the IEEE/CVF Conference on Computer Vision
  and Pattern Recognition ({CVPR})}, pp.\  3626--3636, 2022.

\bibitem[Smaira et~al.(2020)Smaira, Carreira, Noland, Clancy, Wu, and
  Zisserman]{smaira2020short}
Lucas Smaira, Jo{\~a}o Carreira, Eric Noland, Ellen Clancy, Amy Wu, and Andrew
  Zisserman.
\newblock A short note on the kinetics-700-2020 human action dataset.
\newblock \emph{arXiv preprint arXiv:2010.10864}, 2020.

\bibitem[Song et~al.(2021)Song, Meng, and Ermon]{song2021denoising}
Jiaming Song, Chenlin Meng, and Stefano Ermon.
\newblock Denoising diffusion implicit models.
\newblock In \emph{International Conference on Learning Representations
  ({ICLR})}, 2021.

\bibitem[Soomro et~al.(2012)Soomro, Zamir, and Shah]{soomro2012ucf101}
Khurram Soomro, Amir~Roshan Zamir, and Mubarak Shah.
\newblock {UCF101}: A dataset of 101 human actions classes from videos in the
  wild.
\newblock \emph{arXiv preprint arXiv:1212.0402}, 2012.

\bibitem[Srivastava et~al.(2015)Srivastava, Mansimov, and
  Salakhudinov]{srivastava2015unsupervised}
Nitish Srivastava, Elman Mansimov, and Ruslan Salakhudinov.
\newblock Unsupervised learning of video representations using lstms.
\newblock In \emph{International Conference on Machine Learning ({ICML})}, pp.\
   843--852. PMLR, 2015.

\bibitem[Tomar(2006)]{tomar2006converting}
Suramya Tomar.
\newblock Converting video formats with ffmpeg.
\newblock \emph{Linux Journal}, 2006\penalty0 (146):\penalty0 10, 2006.

\bibitem[Tran et~al.(2015)Tran, Bourdev, Fergus, Torresani, and
  Paluri]{tran2015learning}
Du~Tran, Lubomir Bourdev, Rob Fergus, Lorenzo Torresani, and Manohar Paluri.
\newblock Learning spatiotemporal features with 3d convolutional networks.
\newblock In \emph{Proceedings of the IEEE International Conference on Computer
  Vision ({ICCV})}, pp.\  4489--4497, 2015.

\bibitem[Tulyakov et~al.(2018)Tulyakov, Liu, Yang, and
  Kautz]{tulyakov2018mocogan}
Sergey Tulyakov, Ming-Yu Liu, Xiaodong Yang, and Jan Kautz.
\newblock Mocogan: Decomposing motion and content for video generation.
\newblock In \emph{Proceedings of the IEEE Conference on Computer Vision and
  Pattern Recognition ({CVPR})}, pp.\  1526--1535, 2018.

\bibitem[Unterthiner et~al.(2019)Unterthiner, van Steenkiste, Kurach, Marinier,
  Michalski, and Gelly]{unterthiner2019fvd}
Thomas Unterthiner, Sjoerd van Steenkiste, Karol Kurach, Rapha{\"{e}}l
  Marinier, Marcin Michalski, and Sylvain Gelly.
\newblock {FVD:} {A} new metric for video generation.
\newblock In \emph{Deep Generative Models for Highly Structured Data, {ICLR}
  2019 Workshop}, 2019.

\bibitem[Van Den~Oord et~al.(2016)Van Den~Oord, Kalchbrenner, and
  Kavukcuoglu]{van2016pixel}
A{\"a}ron Van Den~Oord, Nal Kalchbrenner, and Koray Kavukcuoglu.
\newblock Pixel recurrent neural networks.
\newblock In \emph{International Conference on Machine Learning ({ICML})}, pp.\
   1747--1756. PMLR, 2016.

\bibitem[Vaswani et~al.(2017)Vaswani, Shazeer, Parmar, Uszkoreit, Jones, Gomez,
  Kaiser, and Polosukhin]{vaswani2017attention}
Ashish Vaswani, Noam Shazeer, Niki Parmar, Jakob Uszkoreit, Llion Jones,
  Aidan~N Gomez, {\L}ukasz Kaiser, and Illia Polosukhin.
\newblock Attention is all you need.
\newblock \emph{Advances in Neural Information Processing Systems ({NeurIPS})},
  30, 2017.

\bibitem[Voleti et~al.(2022)Voleti, Jolicoeur-Martineau, and
  Pal]{voleti2022mcvd}
Vikram Voleti, Alexia Jolicoeur-Martineau, and Chris Pal.
\newblock Mcvd-masked conditional video diffusion for prediction, generation,
  and interpolation.
\newblock \emph{Advances in Neural Information Processing Systems ({NeurIPS})},
  35:\penalty0 23371--23385, 2022.

\bibitem[Vondrick et~al.(2016)Vondrick, Pirsiavash, and
  Torralba]{vondrick2016generating}
Carl Vondrick, Hamed Pirsiavash, and Antonio Torralba.
\newblock Generating videos with scene dynamics.
\newblock \emph{Advances in Neural Information Processing Systems ({NeurIPS})},
  29, 2016.

\bibitem[Wang et~al.(2023{\natexlab{a}})Wang, Yuan, Chen, Zhang, Wang, and
  Zhang]{wang2023modelscope}
Jiuniu Wang, Hangjie Yuan, Dayou Chen, Yingya Zhang, Xiang Wang, and Shiwei
  Zhang.
\newblock Modelscope text-to-video technical report.
\newblock \emph{arXiv preprint arXiv:2308.06571}, 2023{\natexlab{a}}.

\bibitem[Wang et~al.(2023{\natexlab{b}})Wang, Yang, Tuo, He, Zhu, Fu, and
  Liu]{wang2023videofactory}
Wenjing Wang, Huan Yang, Zixi Tuo, Huiguo He, Junchen Zhu, Jianlong Fu, and
  Jiaying Liu.
\newblock Videofactory: Swap attention in spatiotemporal diffusions for
  text-to-video generation.
\newblock \emph{arXiv preprint arXiv:2305.10874}, 2023{\natexlab{b}}.

\bibitem[Wang et~al.(2019)Wang, Wu, Chen, Li, Wang, and Wang]{wang2019vatex}
Xin Wang, Jiawei Wu, Junkun Chen, Lei Li, Yuan-Fang Wang, and William~Yang
  Wang.
\newblock Vatex: A large-scale, high-quality multilingual dataset for
  video-and-language research.
\newblock In \emph{Proceedings of the IEEE/CVF International Conference on
  Computer Vision ({ICCV})}, pp.\  4581--4591, 2019.

\bibitem[Wu et~al.(2022)Wu, Liang, Ji, Yang, Fang, Jiang, and Duan]{wu2022nuwa}
Chenfei Wu, Jian Liang, Lei Ji, Fan Yang, Yuejian Fang, Daxin Jiang, and Nan
  Duan.
\newblock N{\"u}wa: Visual synthesis pre-training for neural visual world
  creation.
\newblock In \emph{European Conference on Computer Vision ({ECCV})}, pp.\
  720--736. Springer, 2022.

\bibitem[Xu et~al.(2018)Xu, Zhang, Huang, Zhang, Gan, Huang, and
  He]{xu2018attngan}
Tao Xu, Pengchuan Zhang, Qiuyuan Huang, Han Zhang, Zhe Gan, Xiaolei Huang, and
  Xiaodong He.
\newblock Attngan: Fine-grained text to image generation with attentional
  generative adversarial networks.
\newblock In \emph{Proceedings of the IEEE Conference on Computer Vision and
  Pattern Recognition ({CVPR})}, pp.\  1316--1324, 2018.

\bibitem[Yan et~al.(2021)Yan, Zhang, Abbeel, and Srinivas]{yan2021videogpt}
Wilson Yan, Yunzhi Zhang, Pieter Abbeel, and Aravind Srinivas.
\newblock Videogpt: Video generation using vq-vae and transformers.
\newblock \emph{arXiv preprint arXiv:2104.10157}, 2021.

\bibitem[Yu et~al.(2022)Yu, Xu, Koh, Luong, Baid, Wang, Vasudevan, Ku, Yang,
  Ayan, Hutchinson, Han, Parekh, Li, Zhang, Baldridge, and Wu]{yu2022scaling}
Jiahui Yu, Yuanzhong Xu, Jing~Yu Koh, Thang Luong, Gunjan Baid, Zirui Wang,
  Vijay Vasudevan, Alexander Ku, Yinfei Yang, Burcu~Karagol Ayan, Ben
  Hutchinson, Wei Han, Zarana Parekh, Xin Li, Han Zhang, Jason Baldridge, and
  Yonghui Wu.
\newblock Scaling autoregressive models for content-rich text-to-image
  generation.
\newblock \emph{Transactions on Machine Learning Research}, 2022, 2022.

\bibitem[Zhang et~al.(2023)Zhang, Zhang, Zhang, and
  Kweon]{zhang2023texttoimage}
Chenshuang Zhang, Chaoning Zhang, Mengchun Zhang, and In~So Kweon.
\newblock Text-to-image diffusion model in generative ai: A survey.
\newblock \emph{arXiv preprint arXiv:2303.07909}, 2023.

\bibitem[Zhang et~al.(2021)Zhang, Wu, Weng, Fu, Chen, Jiang, and
  Davis]{zhang2021videolt}
Xing Zhang, Zuxuan Wu, Zejia Weng, Huazhu Fu, Jingjing Chen, Yu-Gang Jiang, and
  Larry~S Davis.
\newblock Videolt: Large-scale long-tailed video recognition.
\newblock In \emph{Proceedings of the IEEE/CVF International Conference on
  Computer Vision ({ICCV})}, pp.\  7960--7969, 2021.

\bibitem[Zhou et~al.(2022)Zhou, Wang, Yan, Lv, Zhu, and
  Feng]{zhou2022magicvideo}
Daquan Zhou, Weimin Wang, Hanshu Yan, Weiwei Lv, Yizhe Zhu, and Jiashi Feng.
\newblock Magicvideo: Efficient video generation with latent diffusion models.
\newblock \emph{arXiv preprint arXiv:2211.11018}, 2022.

\bibitem[Zhu et~al.(2023)Zhu, Chen, Shen, Li, and Elhoseiny]{zhu2023minigpt}
Deyao Zhu, Jun Chen, Xiaoqian Shen, Xiang Li, and Mohamed Elhoseiny.
\newblock {MiniGPT-4}: Enhancing vision-language understanding with advanced
  large language models.
\newblock \emph{arXiv preprint arXiv:2304.10592}, 2023.

\end{thebibliography}


\end{document}